\documentclass[twoside, web, 10pt]{ieeecolor}   %, draft
\usepackage{tmi}
\usepackage{cite}
\usepackage{amsmath,amssymb,amsfonts}
\usepackage{algorithmic}
\usepackage{graphicx}
\usepackage{textcomp}

\usepackage{multirow}
\usepackage{arydshln}

\newcommand{\xc}{{x}^{c}}
\newcommand{\xm}{{x}^{m}}
\newcommand{\xp}{{x}^{pseudo}}
\newcommand{\rmo}{{R}^{m}}

\newcommand{\zp}{z^+}
\newcommand{\zn}{z^-}
\newcommand{\bi}{\mathbf{I}}
\newcommand{\bz}{\mathbf{0}}

\graphicspath{{./figures/}}

\usepackage[colorlinks=true,
			linkcolor=blue,
			citecolor=blue,
			urlcolor=blue]{hyperref}

\usepackage{orcidlink}

\begin{document}

\title{Motion Artifact-Aware Self-Supervised Representation Learning for 3D Brain MRI Motion Artifact Reduction}

\author{Mojtaba Safari\textsuperscript{\orcidlink{0000-0003-3295-328X}}, Shansong Wang, Zach Eidex, Matthew Goette, Tonghe Wang, Zhen Tian, and Xiaofeng Yang\textsuperscript{\orcidlink{0000-0001-9023-5855}}, \IEEEmembership{Member, IEEE}
	\thanks{This research is supported in part by the National Institutes of Health under Award Numbers R01CA272991, R01DE033512 and P30CA008748.   Corresponding author: Xiaofeng Yang (email: {xfyang@uchicago.edu}).}
	\thanks{Mojtaba Safari, Shansong Wang, Zhen Tian, and Xiaofeng Yang are with the Department of Radiation and Cellular Oncology, The University of Chicago, Chicago, IL, USA.}
	\thanks{Zach Eidex, Matthew Goette, and Xiaofeng Yang are with the Radiation Oncology and Winship Cancer Institute, Emory University, Atlanta, GA, USA. Xiaofeng Yang is also with the Wallace H. Coulter Department of Biomedical Engineering, Georgia Institute of Technology and Emory University, Atlanta, GA, USA. }
	\thanks{Tonghe Wang is with the Department of Medical Physics, Memorial Sloan Kettering Cancer Center, New York, NY, USA.	}
	}

\maketitle

\begin{abstract}
	Patient motion remains a source of image degradation in brain MRI, leading to signal loss, blurring, and geometric distortion that compromise quantitative analysis. Existing deep learning methods for motion correction typically rely on paired clean-corrupted data or k-space acquisitions, which are rarely available in clinical settings. We propose SSRL-MAR, a motion artifact-aware unpaired representation learning framework for motion artifact reduction that requires neither paired training data nor explicit motion labels. SSRL-MAR employed a three-stage training strategy: (1) contrastive learning on 3D patches to extract motion representations by contrasting clean and synthetically corrupted images, (2) a motion artifact-aware synthesis network to generate motion artifacts from clean scans, and (3) a motion artifact-aware generator to restore clean volumes using the learned degrader for self-supervised supervision. On \textit{in-silico} dataset, SSRL-MAR achieved PSNR 23.81dB, SSIM 91.55\%, and NMSE 0.79\%. On \textit{in-vivo} MR-ART dataset, the pretrained model reduced motion distortion, and unsupervised domain adaptation further improved anatomical fidelity. Against a source-only supervised model trained on the same simulated pairs, SSRL-MAR improved PSNR by up to 2.0 dB on MR-ART after unsupervised domain adaptation, and remained within 0.25--0.47 dB of an oracle supervised model that requires real paired data unavailable in practice. At the milder motion level, volumetric error in structures such as the corpus callosum and ventricular system decreased by more than 50\%, confirming improved neuroanatomical consistency. These results indicate that SSRL-MAR provides a robust and scalable image-domain solution for 3D brain MRI motion correction, enabling reliable structural quantification in large-scale neuroimaging studies without requiring prospectively acquired pairs or acquisition-specific calibration.

\end{abstract}


\section{Introduction}\label{sec:introduction}

\IEEEPARstart{M}{agnetic} resonance imaging (MRI) is a cornerstone of neuroimaging, offering excellent soft‐tissue contrast and high spatial resolution. However, the intrinsic sensitivity of MRI to patient motion remains a persistent challenge~\cite{https://doi.org/10.1002/jmri.24850}, particularly for high‐resolution scans with long acquisition times. Even small involuntary head movements perturb the k‐space sampling trajectory, leading to ghosting, blurring, and geometric distortions that can substantially degrade image quality and compromise downstream quantitative and qualitative analyses~\cite{10285512, safari2025systematicreviewmetaanalysisaidriven}. The ability to effectively correct for motion artifacts is therefore essential to ensure accurate morphometric measurements and reliable biomarkers in both research and clinical MRI studies.

Motion artifact reduction (MAR) strategies are typically categorized as prospective or retrospective. Prospective approaches aim to mitigate motion during image acquisition, either through navigator‐based tracking or self‐navigation schemes that estimate motion in real time~\cite{fu2020deep}. Despite their utility, these methods often require hardware modifications, external tracking sensors, or complex sequence adjustments, which can restrict their integration into standard clinical protocols~\cite{https://doi.org/10.1002/mrm.28991}. Moreover, prospective methods are less effective for non‐rigid motion or when fast, flexible imaging is desired. In contrast, retrospective methods operate on the reconstructed images or raw k‐space data to restore quality after acquisition. Classical retrospective approaches include autofocusing and image‐metric optimization~\cite{LIN2006751}, compressed sensing and sparsity‐regularized reconstruction~\cite{https://doi.org/10.1002/mrm.24463}, and model‐based joint estimation of motion and image content~\cite{8252880}. While these methods improved robustness compared with prospective techniques, their computational burden and reliance on explicit motion modeling have limited their adoption for large neuroimaging datasets.

In recent years, deep learning (DL) has emerged as a powerful alternative for retrospective MAR. Supervised models, particularly U-Net‐based architectures, have shown strong artifact‐suppression performance in brain MRI~\cite{DUFFY2021117756, https://doi.org/10.1002/mrm.29255,https://doi.org/10.1002/mrm.27783,https://doi.org/10.1002/mrm.29188}. However, their dependency on paired motion‐corrupted and motion‐free data severely restricts scalability, since such pairs are rarely available in clinical or population‐scale studies. Moreover, supervised models may overfit to specific motion patterns or scanner characteristics, leading to poor generalization.

To overcome the dependence on paired supervision, unsupervised and self‐supervised learning approaches have gained traction. Generative adversarial networks (GANs) and related domain‐translation frameworks~\cite{Liu2021_nature_MI} have been explored to map motion‐corrupted images to their clean counterparts without paired data. While effective at reducing ghosting and blur, adversarial training can be unstable and may hallucinate fine anatomical details, which is an issue for neuroimaging applications that require morphometric precision. Furthermore, some of these methods still require auxiliary clean MR sequences of the same patient for training or domain regularization~\cite{Safari_2024_maudgan}.

DL has also been extended to use the raw k-space, where motion artifacts originate, to remove the motion artifacts. By leveraging acquisition physics, these models can enforce data consistency and achieve physically plausible reconstructions~\cite{spieker2024self, eichhorn2024physics}. However, access to raw k-space is rarely available in large retrospective cohorts, and motion patterns in brain MRI are highly irregular and nonperiodic, posing challenges for model generalization. As a result, there remains a critical need for frameworks that can perform robust motion correction directly in the image domain, without requiring paired data or raw k-space, and that can generalize across motion types and imaging sites.

To address these challenges, we propose SSRL-MAR (Self-Supervised Representation Learning for Motion Artifact Reduction), a motion-artifact–aware framework that extends recent self-supervised degradation-representation learning principles to the domain of 3D MRI motion correction. SSRL-MAR operates entirely in the image domain through a three-stage pipeline that learns to extract, simulate, and correct motion artifacts without paired data. In the first stage, a patch-wise contrastive encoder is trained to learn motion-sensitive representations by contrasting clean and motion-corrupted patches, leading to a latent space that disentangles anatomical content from motion patterns. The second stage leverages this representation to train a motion degrader network that synthesizes realistic motion by simulating the underlying degradation process in a data-driven manner. In the final stage, a motion-aware generator is trained on the synthesized corrupted–clean pairs produced by the first two stages, forming a closed-loop self-supervised system. This unified design enables joint learning of motion representation, artifact synthesis, and artifact correction without requiring ground-truth motion-free references.

We summarize the main contributions of this work as follows:

\begin{enumerate}
	\item SSRL-MAR, an unpaired image-domain framework that unifies contrastive motion representation learning with learned degradation modeling and motion-conditioned correction for 3D brain MRI.
	
	\item A novel motion degrader network that generates realistic motion artifacts conditioned on learned motion embeddings, enabling data-driven unpaired training.
	
	\item Extensive validation across \textit{in-silico} (IXI, HCP) and \textit{in-vivo} (MR-ART) datasets, including comparison with two supervised 3D reference models (oracle and source-only), robustness analysis under both matched and zero-shot motion regimes, and FreeSurfer-based cortical and volumetric evaluation demonstrating improved anatomical fidelity.
	
	\item Comprehensive ablation and representation analyses that quantify the contribution of each component, validate the anatomy-invariant nature of the learned motion embedding, and show their combined impact on stability, fidelity, and cross-domain generalization.
\end{enumerate}

\section{Related Work}\label{sec:related_work}

This section overviews DL approaches for MAR in MRI, organized by training strategy into supervised and unsupervised/self-supervised paradigms. We focus on retrospective correction, where models operate directly on motion-corrupted data to restore diagnostic quality. Approaches span image-domain networks, generative frameworks, and physics-informed models, reflecting diverse strategies to mitigate motion without modifying MRI hardware or acquisition protocols.

\subsection{Supervised Methods}

Supervised MAR relies on paired motion-corrupted and motion-free data to learn direct mappings from degraded to clean images. Most techniques adopt U-Net-based backbones, achieving strong performance for brain MRI~\cite{DUFFY2021117756, https://doi.org/10.1002/mrm.29255, https://doi.org/10.1002/mrm.27783, https://doi.org/10.1002/mrm.29188}. Enhancements with residual connections, attention mechanisms, and multi-scale designs further improve artifact suppression and robustness~\cite{Sommer416, https://doi.org/10.1002/mrm.28719, LIU202069_MRM, ALMASNI2022119411}. Beyond image-domain learning, diffusion probabilistic frameworks have outperformed GAN baselines in training stability and anatomical preservation~\cite{10285512, https://doi.org/10.1002/mp.16844}. Acquisition-domain supervised methods embed motion operators into the encoding model for joint image-motion reconstruction~\cite{https://doi.org/10.1002/mrm.24615, https://doi.org/10.1002/mrm.26796, 8252880}, or couple spatial inference with explicit k-space consistency in dual-domain designs~\cite{10.1007/978-3-031-43999-5_28, safari2025physicsinformeddeeplearningmodel}. While these approaches achieve state-of-the-art results when paired targets or raw k-space data are available, their scalability in large neuroimaging cohorts is limited by the scarcity of such data and potential overfitting to site/vendor-specific motion patterns.

\subsection{Unsupervised and Self-supervised Methods}

Unsupervised MAR removes the dependency on motion-free references by learning directly from corrupted or partially corrupted datasets. Early studies posed the task as image-to-image translation with adversarial learning~\cite{Liu2021_nature_MI, Safari_2024_maudgan, KIM2025109978}, effectively reducing ghosting and blur but risking instability and hallucination of fine structures which is problematic for morphometry. More recently, diffusion-based generative models have emerged as a powerful non-adversarial alternative. However, most diffusion applications for MAR still rely on supervised training~\cite{10285512}, and even recent unsupervised adaptations~\cite{SARKAR2025108684} operate by learning to reverse a generic noising process that is not physically grounded in the mechanics of MRI motion.

Parallel to these generative approaches, degradation-representation learning has been explored through unsupervised two-stage pipelines that first learn motion distributions from corrupted data and then synthesize realistic degradations to supervise restoration~\cite{angella2025dimadiffusingmotionartifacts, 10375761}. While this simulation-restoration decoupling alleviates the need for paired data, separating representation learning from correction can weaken domain-invariant motion priors and limit anatomy preservation, as the learned degradation representation is not explicitly optimized for the subsequent restoration task.

Unsupervised learning has also extended into the acquisition domain. Physics-informed and neural implicit representations jointly model anatomy and motion trajectories as continuous functions in k-space~\cite{spieker2024self, wu2025moner}, offering strong data fidelity and physical interpretability. Complementary corrupted-line detection excludes motion-affected k-space samples to improve quantitative mapping~\cite{eichhorn2024physics, https://doi.org/10.1002/mrm.70050, 10639524, 10.1007/978-3-031-72104-5_37}. However, dependence on raw k-space, which is rarely retained in retrospective archives and varies across scanners and vendors, limits broad clinical deployment and cross-site scalability.

Our proposed method, SSRL-MAR, addresses the limitations of these prior approaches through a unified image-domain framework with two key innovations. First, unlike standard degradation-representation learning where the degradation embedding is learned independently, SSRL-MAR introduces a tightly coupled, three-stage co-adaptation. The motion representation learned via contrastive pretraining is explicitly engineered to be anatomy-invariant and is then used to condition \textit{both} the synthesis of artifacts and their removal. This bidirectional conditioning creates a closed-loop system where the representation is simultaneously optimized for generating realistic corruptions and for guiding their correction. Second, and more fundamentally, our framework provides an explicit, interpretable model of the motion process itself. Rather than learning to reverse an implicit noising process as in diffusion models~\cite{SARKAR2025108684}, we introduce a motion-conditioned degrader network that learns to approximate the conditional distribution \(p(\xm|\xc, m)\). This learned simulator, guided by the anatomy-invariant motion embedding, generates realistic artifacts directly in the image domain, and the restoration network learns to invert this \textit{explicitly modeled} degradation. This establishes a physics-inspired, analysis-by-synthesis pathway for artifact removal that is conceptually distinct from the implicit denoising paradigm of generative models. By integrating a contrastive encoder with a learned degrader and a motion-aware generator, SSRL-MAR effectively eliminates the need for paired data or raw k-space while providing a practical and scalable solution that maintains anatomical fidelity, which we demonstrate through extensive validation on both simulated and \textit{in-vivo} datasets.

\section{Methods}\label{sec:methods}

\begin{figure}[t!]
	\centering
	\includegraphics[width=0.48\textwidth, draft=false]{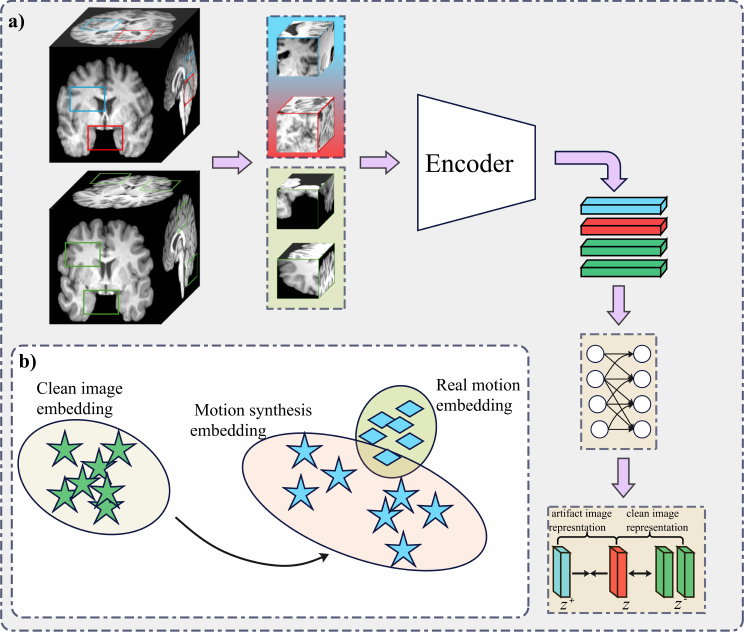}
	\caption{Unsupervised motion artifact representation learning. (a) 3D patches ($96^3$) from clean and motion-corrupted volumes are encoded into 256-D embeddings via contrastive learning. (b) The latent space separates clean embeddings from motion-affected ones (both real and degrader-synthesized), with the contrastive objective aligning artifact embeddings ($z^+$) while separating clean ones ($z^-$).}
	\label{fig:figure1_contrastiveLearning}
\end{figure}

\subsection{Problem Formulation and Overview}

Let $\xc \sim p(\xc)$ denote motion-free 3D brain MRI volumes and $\xm \sim p(\xm)$ denote motion-corrupted volumes acquired from different subjects. The objective is to learn a mapping $\xm \rightarrow \xc$ without access to paired training data. The proposed SSRL-MAR framework addresses this challenge through a three-stage approach that combines contrastive representation learning, degradation modeling, and self-supervised restoration.

In the first stage, motion-sensitive representations are learned through contrastive learning between clean and motion-corrupted image patches. The second stage employs a motion-conditioned degrader network to synthesize realistic motion artifacts from clean images. The final stage trains a generator to remove artifacts using the synthesized pairs for self-supervised supervision. This closed-loop system enables comprehensive motion artifact reduction without requiring paired data or explicit motion labels.

\begin{figure*}[!t]
	\centering
	\includegraphics[width=0.98\textwidth, draft=false]{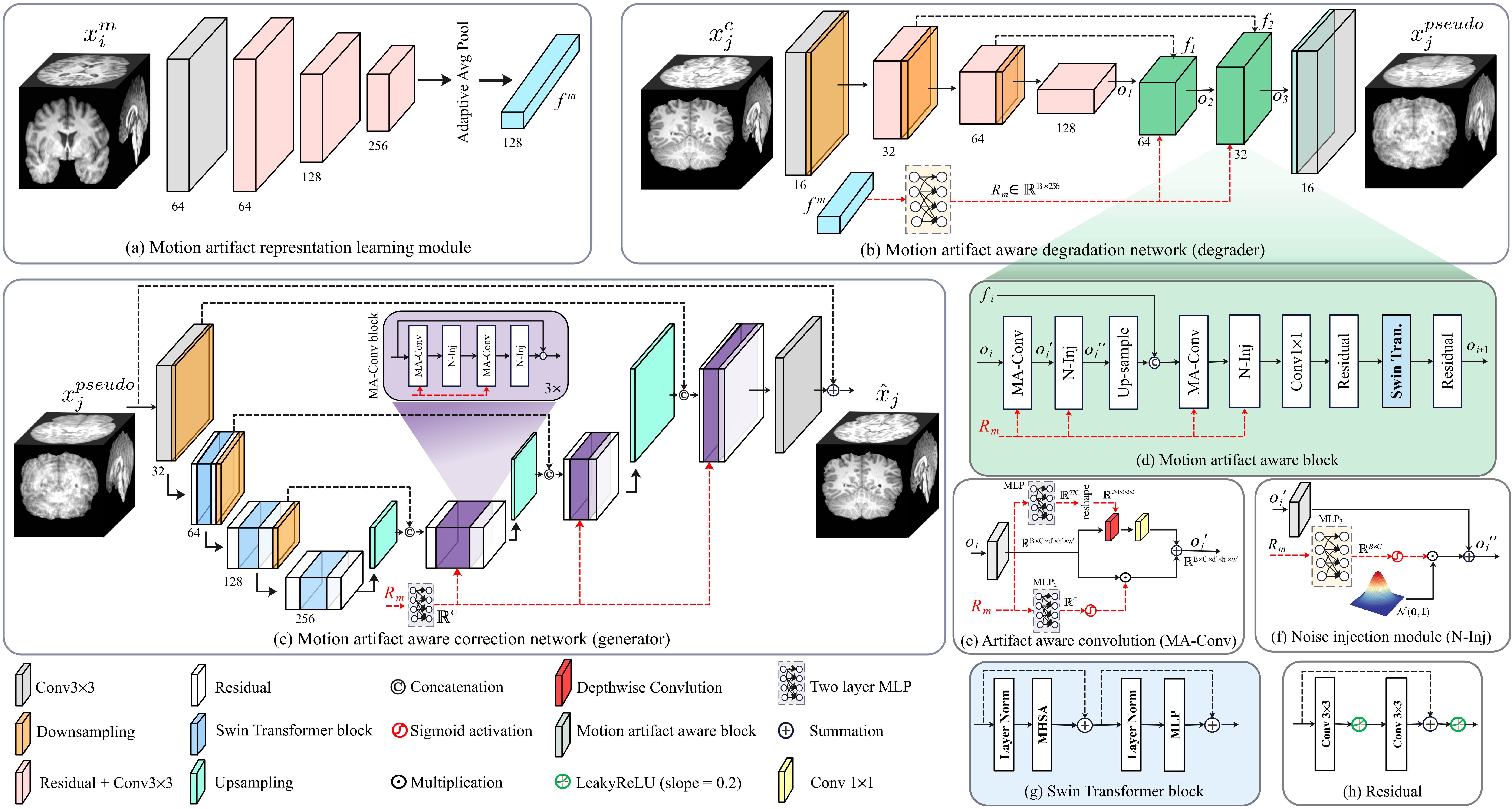}
	\caption{Overview of SSRL-MAR. (a) Contrastive encoder learns motion-sensitive embeddings from clean ($x_j^c$) and motion-corrupted ($x_i^m$) patches, drawn as unpaired samples from different subjects. (b) Motion-conditioned degrader synthesizes pseudo motion-corrupted volumes $x_j^{\mathrm{pseudo}}$ from clean inputs using the learned motion representation $R_m$. (c) Restoration network reconstructs $\hat{x}_j$ from $x_j^{\mathrm{pseudo}}$ using embedded motion priors. (d) Artifact-aware block, (e) Motion-Aware Convolution (MA-Conv), and (f) Noise-Injection module (N-Inj) inject $R_m$ into convolutional pathways. Dashed red arrows denote motion-conditioned modulation.}
	\label{fig:pipeline}
\end{figure*}

\subsection{Stage 1: Motion Artifact Representation Learning}\label{sec:artifact_representation_learning}
\subsubsection{Contrastive Learning Framework}

The objective of this stage is to obtain discriminative embeddings that encode motion-induced degradations. An unsupervised contrastive learning framework~\cite{9157636} is employed using unpaired motion-corrupted and motion-free image patches, as illustrated in Fig.~\ref{fig:figure1_contrastiveLearning}(a). 

For each motion-corrupted patch, two augmented views are created as query and positive pairs. Negative samples are drawn from independent motion-free patches. A shared encoder followed by a two-layer multilayer perceptron projection head maps these patches into 256-dimensional embeddings $z$, $z^{+}$, and $z^{-}$.

The intended effect of this contrastive objective is visualized conceptually in Fig.~\ref{fig:figure1_contrastiveLearning}(b). The learned embedding space separates clean image representations from motion-affected ones, with real motion-corrupted embeddings and those synthesized by our subsequent degrader network forming overlapping distributions that are distinct from the cluster of clean embeddings. This separation enables the disentanglement of motion artifacts from anatomical content.

This design is critical for learning an anatomy-invariant motion representation. Since the negative samples are sourced from anatomically distinct, motion-free volumes of different subjects, the only consistent feature that allows the network to identify the positive pair is the shared motion pattern. The encoder is therefore compelled to become invariant to anatomical content and sensitive to motion-induced distortions to successfully minimize the contrastive loss.

The contrastive encoder employs a 3D ResNet backbone that processes $96^3$ patches randomly cropped from clean and motion-corrupted volumes. The network comprises three residual blocks with channel dimensions progressing from 64 to 256, followed by group normalization and LeakyReLU (negative slope = 0.2) activations throughout the architecture.

\subsubsection{Anisotropic Contrastive Loss}

We employ an anisotropic InfoNCE loss~\cite{rusak2025infonceidentifyinggaptheory} that reweights feature dimensions to emphasize motion-sensitive directions, unlike standard contrastive approaches~\cite{pmlr-v119-chen20j,chen2020improvedbaselinesmomentumcontrastive}. The loss is defined as:
\begin{equation}
	\begin{aligned}
		&\mathcal{L}_{\mathrm{deg}} = -\sum_{{i=1}}^{B}\log\\&\frac{\exp(-d_{\Lambda}(z_i, \zp_i)/\tau)}{\exp(-d_{\Lambda}(z_i, \zp_i)/\tau) + \sum_{n=1}^{N_\mathrm{queue}} \exp(-d_{\Lambda}(z_i, \zn_n)/\tau)},
	\end{aligned}
	\label{eq:aniso_infoNCE}
\end{equation}
where $d_{\Lambda}(z, z') = (z - z')^{\top} \mathrm{diag}(\exp(\alpha)) (z - z')$ is the anisotropic distance, $\alpha \sim \mathcal{N}(\bz, 0.1\bi)$ are learned dimension weights, $\tau=0.07$ is the temperature, and $N_\text{queue}=4096$.

\subsection{Stage 2: Motion Artifact-Aware Synthesis}\label{sec:artifact_aware_motion_synthesis}

\subsubsection{Probabilistic Formulation}

The motion artifact-aware degradation network aims to synthesize motion-corrupted images $\xp_j$ from motion-free inputs $\xc_j \sim p(\xc)$ such that $\xp_j$ resembles samples from the real motion-corrupted distribution $p(\xm)$, where $\xm_i \sim p(\xm)$ and $i \ne j$. 

Formally, our objective is to learn the conditional distribution $p(\xm|\xc)$. Since motion artifacts arise from underlying motion parameters $m \sim p(m)$, we model the degradation process through the factorization:
\begin{equation}
	p(\xm|\xc) = \int p(\xm|\xc, m) p(m)  dm,
\end{equation}

In our framework, the encoder from Stage 1 (Fig.~\ref{fig:pipeline}(a)), parameterized by $\theta_E$, learns to extract motion representations by modeling the posterior distribution $p_{\theta_E}(m | \xm)$ from unpaired motion-corrupted data. The degrader network $\mathcal{D}_\varphi$ subsequently learns to approximate $p(\xm | \xc, m)$ by generating realistic motion artifacts conditioned on both clean images $\xc$ and the learned motion representation $\rmo \sim p_{\theta_E}(m | \xm)$. This approach enables sampling from the target distribution $p(\xm|\xc)$ without requiring paired training data.

\subsubsection{Degrader Network Architecture}

The motion-aware degrader $\mathcal{D}_\varphi$ is constructed using a U-Net backbone (see Fig.~\ref{fig:pipeline}(b))~\cite{10643318}. The encoder pathway begins with the motion-free image $\xc_j \in \mathbb{R}^{1 \times D \times H \times W}$ processed through an initial $3\times3$ convolution and downsampling layer, followed by three residual blocks with channel dimensions of 32, 64, and 128. Each processing stage incorporates $3\times3$ convolution layers with stride 2 for progressive downsampling.

The decoder pathway contains two motion artifact-aware (MAA) blocks which is a combination of motion-aware convolution layer (MA-Conv; see Section~\ref{sec:ma_conv_bloack}) that integrate the motion representation $\rmo$ through specialized convolution layers. In addition to the $\rmo$, each motion artifact-aware block employed skip connections and residual blocks with descending channel dimensions of 128, 64, and 32. The architecture concludes with a final $3\times3$ convolution layer that maps features back to image space.

\subsubsection{Motion-Aware Components}\label{sec:ma_conv_bloack}

The MA-Conv layer, illustrated in Fig.~\ref{fig:pipeline}(e), modulates features based on the learned motion representation. Given feature $O_i$ at layer $i$, the MA-Conv operation computes:

\begin{equation}
	\centering
	\begin{aligned}
		{O}_i^\prime &= {O}_i \odot \sigma\left(\mathrm{MLP}_2(\rmo) + \mathrm{Conv}_{1\times1}\left(\mathrm{DWConv}_{3\times3}\left({O}_i; K_i\right)\right)\right), \\
		K_i &= \mathrm{MLP}_1(\rmo)
	\end{aligned}
	\label{eq:ma_conv}
\end{equation}
where $\sigma(\cdot)$ denotes the sigmoid activation, $\odot$ represents element-wise multiplication, and $\mathrm{DWConv}_{3\times3}$ and $\mathrm{Conv}_{1\times1}$ represent dynamic depthwise and pointwise convolutions, respectively. The motivation behind using the dynamic depthwise convolution is that the convolution kernels learned for different distortion levels may share similar patterns but different statistics~\cite{He_2019_CVPR, 10738507}. The kernel ${K}_i$ is dynamically generated from the motion representation $\rmo$ via $\mathrm{MLP}_1(\cdot)$, enabling motion-aware feature adaptation. The Noise-Injection module (N-Inj) shown in Fig.~\ref{fig:pipeline}(f), which introduces stochasticity into the degradation process, is defined as follows:

\begin{equation}
	\centering
		O_i^{\prime\prime} = O_i^\prime + \Big[ \sigma\Big(\mathrm{MLP}_3(\rmo)\Big) \odot \epsilon \Big],\quad\epsilon \sim \mathcal{N}(\bz, \bi),
	\label{eq:n_inj}
\end{equation}

\subsubsection{Degrader Training Objectives}

The degrader network is optimized using a composite loss function:
\begin{equation}
	\mathcal{L}_D = \lambda_{\mathrm{content}} \mathcal{L}_{\mathrm{content}} + \lambda_{\mathrm{adv}} \mathcal{L}_{\mathrm{adv}} + \lambda_{\mathrm{deg}} \mathcal{L}_{\mathrm{deg}},
	\label{eq:degrader_overall_loss}
\end{equation}
with empirically determined weighting coefficients $\lambda_{\mathrm{content}} = 10$, $\lambda_{\mathrm{adv}} = 0.5$, and $\lambda_{\mathrm{deg}} = 0.005$.

The content loss maintains structural consistency between synthesized and clean images:
\begin{equation}
	\mathcal{L}_{\mathrm{content}} = \parallel \Big( (\xc_j\downarrow) - (\xp_j\downarrow) \Big) \odot M \parallel_1,
	\label{eq:content_loss_v01}
\end{equation}
where $ M $ and $\downarrow$ denote brain mask and a downsampling operation included to reduce computational requirements. The brain mask $M$ ensures the loss focuses exclusively on biologically relevant regions, preventing the model from learning spurious correlations in background or non-brain tissues. The downsampling operation $\downarrow$ is applied to reduce the computational burden of the content loss while preserving its effectiveness in enforcing global structural consistency.

The $\mathcal{L}_{\mathrm{content}}$ plays a vital role in enforcing the disentanglement of motion and anatomy. The degrader must learn to use the motion representation $\rmo$ (extracted from a different subject's scan $\xm_i$) to degrade the clean input $\xc_j$ without altering its underlying anatomical structure. Minimizing $\mathcal{L}_{\mathrm{content}}$ ensures that the output $\xp_j$ is a corrupted version of $\xc_j$, not a morphing of $\xc_j$ towards the anatomy of $\xm_i$. This is only feasible if $\rmo$ is effectively purged of anatomical information and contains a generalized representation of the motion degradation itself.

The adversarial loss ensures synthesized images reside in the motion-corrupted domain using a PatchGAN discriminator~\cite{Isola_2017_CVPR} enhanced with spectral normalization~\cite{miyato2018spectral} for training stability:
\begin{equation}
	\begin{aligned}
		\mathcal{L}_{\mathrm{adv}}^{\mathrm{D}} &= \mathbb{E}_{\xm} \Big[ \parallel \mathcal{D}(\xm) - y_{\mathrm{real}} \parallel_2^2 \Big] \\
		&+ \mathbb{E}_{\xp} \Big[ \parallel \mathcal{D}(\xp) - y_{\mathrm{fake}} \parallel_2^2 \Big] \\
		\mathcal{L}_{\mathrm{adv}}^{\mathrm{G}} &= \mathbb{E}_{\xp} \Big[ \parallel \mathcal{D}(\xp) - 1 \parallel_2^2 \Big],
	\end{aligned}
	\label{eq:lsgan_adv}
\end{equation}
where $y_{\mathrm{real}} \sim \mathcal{U}(0.8, 1.0)$ and $y_{\mathrm{fake}} \sim \mathcal{U}(0.0, 0.2)$ represent smoothed labels that enhance training robustness.

The contrastive loss component ($ \mathcal{L}_{\text{deg}} $) employs the same anisotropic InfoNCE loss defined in \eqref{eq:aniso_infoNCE} to ensure that synthesized artifacts exhibit embedding characteristics similar to real motion-corrupted images.

\subsection{Stage 3: Motion Artifact-Aware Correction}\label{sec:artifact_aware_mar}

\subsubsection{Generator Network Architecture}

The generator $\mathcal{G}_\psi$ utilizes a U-Net backbone architecture, shown in Fig.~\ref{fig:pipeline}(c) designed to remove motion artifacts while preserving essential anatomical details. The network accepts synthesized images $\xp_j \in \mathbb{R}^{1 \times D \times H \times W}$ as input, which first pass through an initial $3\times3$ convolution and downsampling layer to produce feature map $f_1 \in \mathbb{R}^{32 \times \frac{D}{2} \times \frac{H}{2} \times \frac{W}{2}}$.

The encoding pathway consists of three processing layers that alternate between residual blocks and Swin Transformer blocks (see Fig.~\ref{fig:pipeline}(g)), progressively extracting hierarchical features. The decoding pathway employs three corresponding layers that combine residual blocks (see Fig.~\ref{fig:pipeline}(h)) with motion-aware convolution blocks, incorporating the motion representation $\rmo$ through MA-Conv operations. The architecture concludes with a final $3\times3$ convolution layer that generates the motion-corrected output $\hat{x}_j$.

\subsubsection{Generator Training Objective}

The generator optimization employs a composite loss function that combines both intensity and gradient consistency terms:

\begin{equation}
	\begin{aligned}
		\mathcal{L}_{\mathrm{gen}} = &\lambda_{\mathrm{consis}} \parallel (\hat{x}_j - \xc_j) \odot M \parallel_1 + \\
		&\lambda_{\mathrm{grad}}\parallel (\nabla(\hat{x}_j - \xc_j)) \odot M \parallel_1,
	\end{aligned}
	\label{eq:generator_loss}
\end{equation}
where $\lambda_{\mathrm{consis}}=10$ and $\lambda_{\mathrm{grad}}=0.1$ are the empirically determined weighting factors, $\nabla$ denotes the spatial gradient operator, and $M$ represents the brain mask that focuses the optimization on biologically relevant regions

The first term enforces intensity-level consistency between the generated output and the pseudo motion-corrupted input, ensuring overall structural alignment. The second term introduces gradient consistency, which preserves edge information and fine anatomical details by matching the spatial derivatives of the generated and target images. This dual-objective approach enhances the preservation of cortical boundaries and subtle anatomical features that are critical for neuroimaging applications

\subsection{Progressive Training Strategy} \label{sec:progressive_training_strategy}

The training process follows a sequential three-stage approach to ensure stable convergence and effective disentanglement of motion and anatomical features. In the first stage, contrastive pretraining minimizes $\mathcal{L}_{\text{deg}}$ using random $96^3$ patches extracted from unpaired clean and motion-corrupted volumes. This stage executes for 150 epochs with a batch size of 8, producing frozen encoder weights $\theta_E$.

The second stage focuses on degrader training by minimizing the composite loss $\mathcal{L}_D$ while maintaining the encoder in a frozen state. This phase runs for 50 epochs with a reduced batch size of 2, utilizing the AdamW optimizer with a learning rate of $1\times10^{-4}$.

The final stage trains the generator to minimize $\mathcal{L}_{\text{gen}}$ with both encoder and degrader networks frozen. Due to memory constraints associated with processing full 3D volumes, this stage employs a batch size of 1 across 50 training epochs, with optimization performed using AdamW and a learning rate of $5\times10^{-5}$.

All experiments were conducted on an NVIDIA A100 (80GB) GPU using PyTorch version 2.5.1 (built with CUDA 12.1). Training times for each stage were as follows: Stage 1 required 7 hours 31 minutes, Stage 2 required 12 hours 7 minutes, and Stage 3 required 61 hours 24 minutes. Peak GPU memory usage during training was approximately 60 GB for Stage 3 due to full-volume processing, while Stages 1 and 2 operated on patches and required approximately 79 GB. For inference, SSRL-MAR processes a full 3D volume ($168 \times 168 \times 192$ voxels) in approximately 7 seconds on the same GPU, making it suitable for integration into retrospective processing pipelines.

\section{EXPERIMENTS AND RESULTS}\label{sec:experiments_results}

\subsection{Datasets}

This study employed three publicly available T1-weighted (T1w) magnitude brain MRI datasets for training and validation: IXI (581 volumes), Human Connectome Project (HCP; 1,113 volumes)~\cite{VANESSEN201362}, and MR-ART (148 volumes)~\cite{Narai2022_mrART}. The IXI and HCP datasets were used to simulate motion-corrupted images for model training and validation, whereas the \textit{in-vivo} MR-ART dataset was used exclusively for external validation.

Motion artifacts were synthetically generated using the \texttt{RandomMotion} transform from TorchIO~\cite{perez_garcia_torchio_2021}. Each simulated volume included 10 random motion events (a setting we maintain across all experiments, including the extreme motion analysis in Section~\ref{sec:robustnessExtremeMotionSeverity}), with rotation angles $\Theta_{r} \sim \mathcal{U}(-2, 2)$ degrees and translations $\Theta_{t} \sim \mathcal{U}(-2, 2)$ mm per event for the training range. For robustness evaluation, the same 10-event protocol was used with extended ranges of $\pm3$, $\pm4$, $\pm5$, $\pm6$, and $\pm7$ degrees and millimeters. These ranges were used both for zero-shot testing of the $\pm2$-trained model and, for $\pm3$, $\pm5$, and $\pm7$, to generate matched training pools for the per-regime analysis in Section~\ref{sec:robustnessExtremeMotionSeverity}.

All images were skull-stripped using HD-BET~\cite{https://doi.org/10.1002/hbm.24750}, followed by center cropping and resampling to a standardized resolution of \(168 \times 168 \times 192\) voxels to remove non-brain regions and ensure consistent input dimensions across datasets. Resampling was performed using linear interpolation, which preserves high-frequency information while avoiding the overshoot artifacts associated with higher-order methods. For the \textit{in-silico} datasets, motion simulation was performed after resampling, ensuring that any interpolation effects are consistent across clean and corrupted pairs. For the \textit{in-vivo} MR-ART dataset, resampling was applied to images containing real motion artifacts; visual inspection confirmed that characteristic motion patterns (ghosting, blurring, ringing) remained clearly distinguishable after interpolation.

The IXI and HCP datasets were randomly divided into training (75\%) and validation (25\%) subsets. The same 75:25 ratio was applied to the MR-ART dataset for unsupervised domain adaptation and external validation.

\subsection{Evaluation}

The proposed SSRL-MAR framework was evaluated using comprehensive quantitative, qualitative, and anatomy-level analyses across both \textit{in-silico} and \textit{in-vivo} datasets. Three representative benchmark methods were included for comparison: a 3D implementation of CycleGAN for unpaired motion-to-clean translation, AutoDPS~\cite{SARKAR2025108684} in its original 2D diffusion-based motion correction formulation, and UDDN~\cite{WU2023107373} in its native 2D framework for unsupervised motion disentanglement. All comparative models were executed using their publicly released implementations, with dimensional adaptations applied where necessary to enable processing of 3D brain MRI volumes.

For UDDN and AutoDPS, which are designed as 2D models, we processed each 3D volume slice-by-slice along the axial plane. Each slice was independently normalized to the range $[-1,+1]$ using the same intensity normalization applied to our 3D method, then passed through the pretrained 2D model. The output slices were reassembled into a 3D volume, and the original volume-level intensity statistics were reapplied to ensure consistent quantitative comparisons. This slice-wise aggregation, while standard, cannot model through-plane motion artifacts or leverage 3D contextual information, which inherently disadvantages these methods relative to native 3D approaches.

To contextualize performance, we trained two supervised variants of our generator using the same architecture. The first, denoted Supervised (oracle), was pretrained on simulated \textit{in-silico} pairs and then fine-tuned on paired real MR-ART scans at each motion level. It serves only as an upper bound, because paired real motion-free and motion-corrupted scans are unavailable in routine practice. The second, denoted Supervised (source-only), was trained solely on the simulated \textit{in-silico} pairs and applied to the in-vivo data without fine-tuning, matching the data regime of our self-supervised method. Both are reported in Table~\ref{tab:image_metrics_qualitative}; the oracle variant is also shown qualitatively in Fig.~\ref{fig:qualitative_fig01}.

To further assess anatomical fidelity beyond voxelwise similarity metrics, volumetric and surface-based segmentation analyses were conducted on 20 randomly selected \textit{in-vivo} MR-ART subjects using FreeSurfer (version 8.1.0)~\cite{FISCHL2012774}. These analyses enabled evaluation of cortical topology preservation, regional boundary consistency, and motion-induced volumetric errors following reconstruction.

\subsubsection{Qualitative results} 

Representative motion-corrupted, motion-corrected, and motion-free images from all datasets are shown to illustrate visual performance. For SSRL-MAR, additional visualizations of volumetric segmentation results and atlas-based cortical parcellations (Desikan-Killiany and Destrieux) were generated to examine gyri-sulci continuity and confirm preservation of fine cortical geometry. These examples highlight the ability of the proposed framework to suppress ghosting, ringing, and blurring while maintaining anatomically plausible structures across diverse motion levels.

\subsubsection{Quantitative results} 

Quantitative assessment was performed using peak signal-to-noise ratio (PSNR), structural similarity index measure (SSIM), and normalized mean-squared error (NMSE) computed within the brain tissue. In addition, FreeSurfer-derived regional volumes were compared with motion-free references to quantify anatomy-level consistency, particularly in motion-sensitive regions such as the corpus callosum and ventricular system.

\begin{figure}[t!] % or [b] or [!ht]
	\centering
	\includegraphics[width=0.48\textwidth, draft=false]{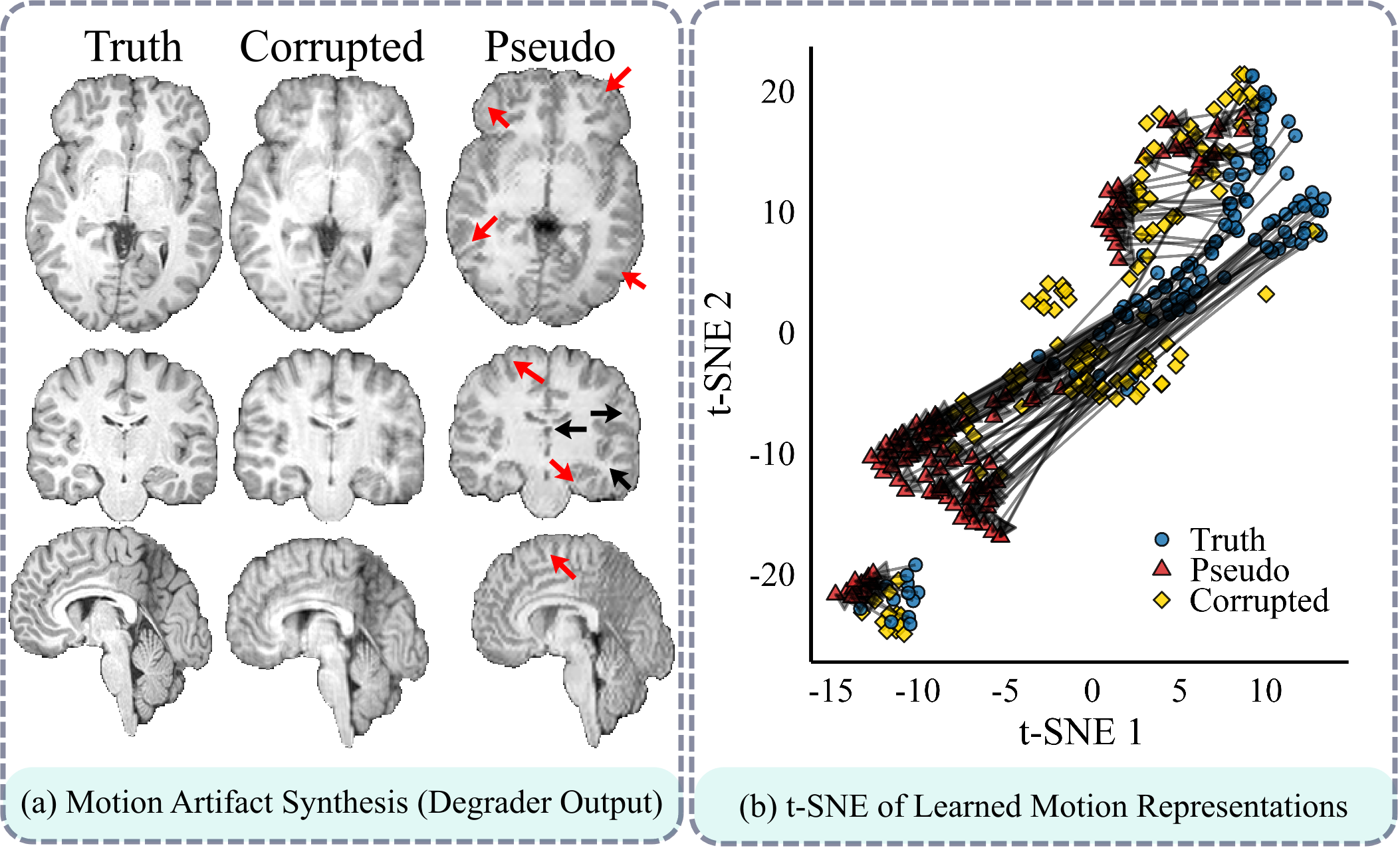}
	\caption{Panel (a) illustrates the degrader network synthesizing realistic motion artifacts from clean inputs without paired supervision, introducing blur (black arrows) and ringing (red arrows) patterns that mimic real corruption. Panel (b) shows the t-SNE embedding of the learned latent space, where motion-free, pseudo-corrupted, and real-corrupted samples form distinct clusters. The degrader mechanism encourages clean samples to shift toward corrupted manifolds, promoting contrastive alignment and enhancing motion-awareness in the learned representations.}
	\label{fig:tsne_01}
\end{figure}

\subsection{Experimental Results}\label{sec:results}
This section presents a comprehensive evaluation of SSRL-MAR for brain motion artifact reduction across multi-center \textit{in-silico} and \textit{in-vivo} datasets. We first analyze the learned motion representation and degradation mechanism, followed by quantitative and qualitative comparisons with benchmark methods. Ablation studies further isolate the contribution of each module, and segmentation-based analyses confirm preservation of neuroanatomical fidelity.

\subsubsection{Motion Representation and Simulation Learning}\label{sec:results_representation}

Fig.~\ref{fig:tsne_01}(a) illustrates an example from the \textit{in-vivo} dataset comparing a clean reference with the degrader's synthesized output. Red arrows indicate realistic motion-induced artifacts, including blur and ringing. The t-SNE visualization in Fig.~\ref{fig:tsne_01}(b) of Stage 1 encoder features shows pseudo-motion features clustering closely with real motion-corrupted features, confirming that the learned representation captures physically meaningful degradation patterns. During Stage 2, the degrader progressively shifts clean embeddings toward the corrupted manifold, validating the effectiveness of the proposed motion simulation strategy.

A slight distribution shift between pseudo-corrupted and real motion-corrupted embeddings is visible in Fig.~\ref{fig:tsne_01}(b), reflecting the inherent difficulty of perfectly simulating the full variability of real motion artifacts. This minor gap does not impair downstream performance, as the generator learns to correct artifact patterns rather than to match specific embedding coordinates. The strong generalization to \textit{in-vivo} MR-ART data (Fig.~\ref{fig:qualitative_fig01}, Table~\ref{tab:image_metrics_qualitative}) confirms that the pseudo-corrupted images capture the essential characteristics necessary for effective motion correction.

We computed silhouette scores to compare clustering of the latent embeddings by subject identity and by motion severity. Using subject identity as cluster labels, the silhouette score was $-0.455$, indicating that embeddings from the same subject do not cluster together in latent space. Values near zero indicate no meaningful clustering structure, whereas negative values indicate that samples are closer to points from other clusters than to samples within the same cluster. 

For comparison, using three motion severity groups defined by SSIM relative to the clean reference, the silhouette score was $-0.020$. Although this value indicates weak clustering, it is substantially higher than that obtained for subject identity, suggesting that the latent space is much less organized by subject-specific anatomy than by motion-related variation.

\begin{figure}[t!]
	\centering
	\includegraphics[width=0.48\textwidth]{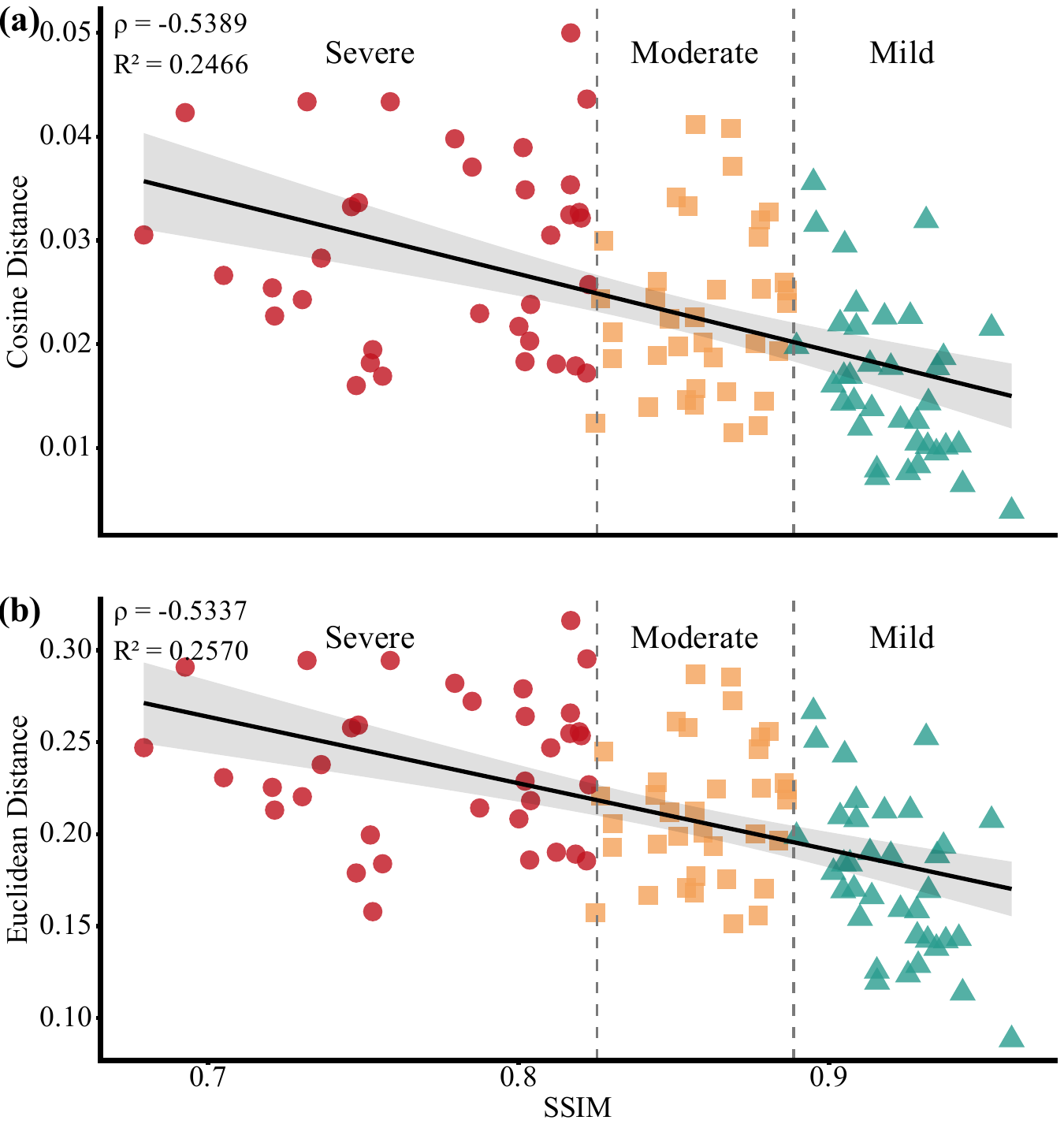} %panel_ab_cosine_euclidean_vs_ssim_to_clean     panel_ab_vertical_cosine_euclidean_vs_ssim_to_clean
	\caption{Latent space analysis. (a) Cosine distance and (b) Euclidean distance between corrupted and clean embeddings versus SSIM. Each point represents one corrupted volume. Vertical dashed lines indicate tertile boundaries ($\mathrm{SSIM}=0.8253$ and $0.8885$) used to define mild, moderate, and severe groups. Negative Spearman correlations ($\rho=-0.53$ and $-0.54$) indicate that latent distance increases as structural similarity decreases. Silhouette scores further show substantially weaker organization by subject identity ($-0.455$) than by motion severity ($-0.020$).}
	\label{fig:latent_monotonicity}
\end{figure}

\begin{figure*}[h!] 
	\centering
	\includegraphics[width=0.98\textwidth, draft=false]{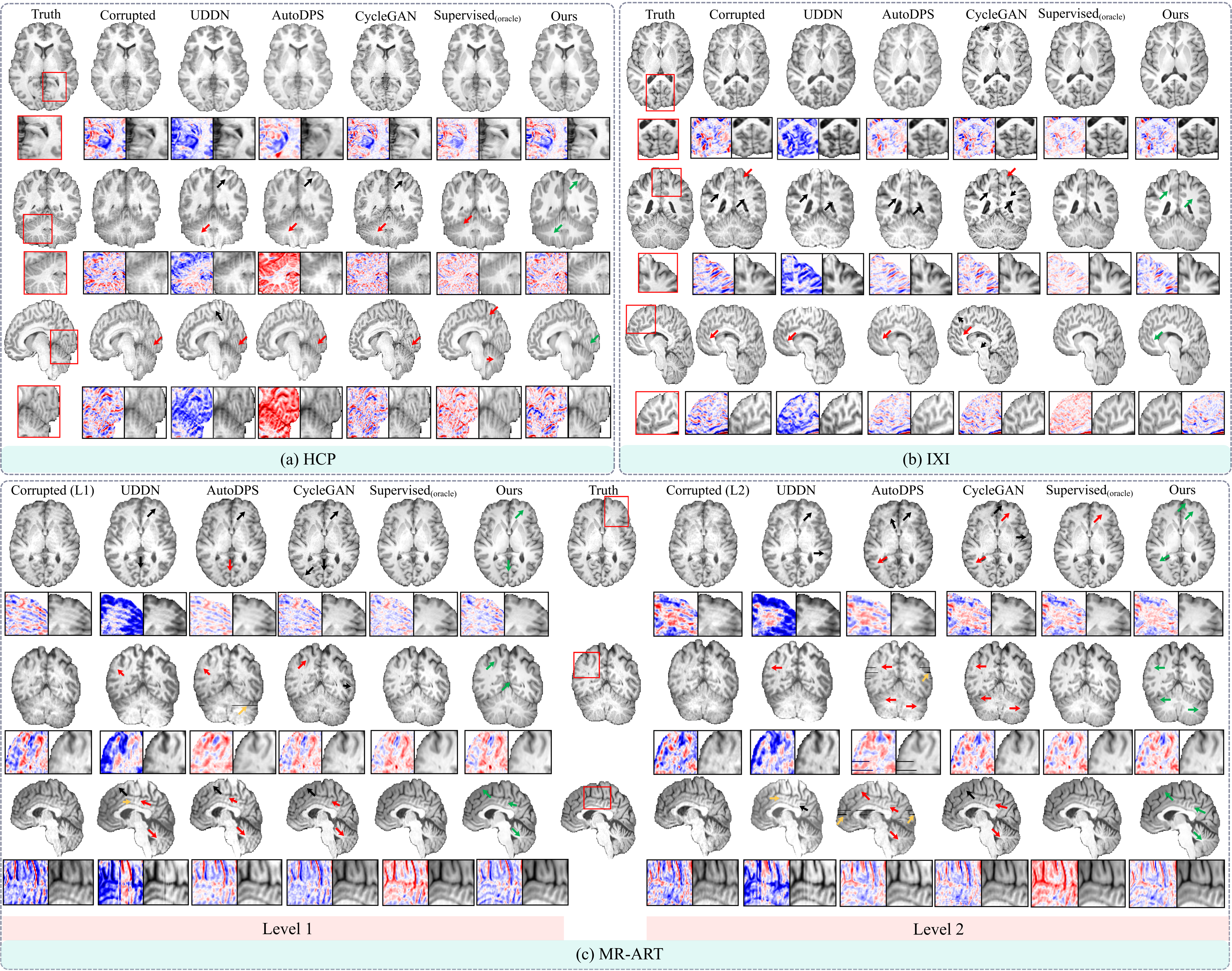}%qualitative_02a.png
	\caption{Qualitative comparison of motion correction methods across three datasets: (a) HCP, (b) IXI, and (c) MR-ART. From left to right: motion-corrupted input, UDDN, AutoDPS, CycleGAN, supervised baseline (oracle, fine-tuned on paired real data), SSRL-MAR (ours), and motion-free reference. Red arrows indicate regions where comparative methods fail to preserve anatomical detail. Black arrows indicate residual motion artifacts. Gold arrows denote slice-wise artifacts introduced by 2D-based models. Green arrows highlight anatomical structures successfully preserved by SSRL-MAR. The supervised baseline, representing an upper bound that requires paired real data unavailable in practice, shows marginally sharper reconstructions than our unsupervised method, consistent with the quantitative results in Table~\ref{tab:image_metrics_qualitative}. However, SSRL-MAR achieves comparable performance without requiring paired training data, making it applicable to real clinical scenarios where such pairs are unavailable.}
	\label{fig:qualitative_fig01}
\end{figure*}

We further analyzed the relationship between latent displacement and motion severity by measuring the distance between each corrupted embedding and its clean reference embedding. As shown in Fig.~\ref{fig:latent_monotonicity}, both Euclidean and cosine distances increase monotonically as SSIM decreases. Spearman rank correlations were $-0.53$ and $-0.54$ for Euclidean and cosine distances, respectively, indicating a strong monotonic association between latent displacement and motion-induced image degradation. Linear regression likewise produced negative slopes with narrow confidence intervals.

Taken together, these analyses indicate that the learned representation suppresses subject-specific anatomical information while remaining sensitive to motion-related distortions.

\subsubsection{Qualitative Motion Correction Evaluation}\label{sec:results_qualitative}
Building on this learned motion representation, Fig.~\ref{fig:qualitative_fig01} demonstrates motion correction performance across \textit{in-silico} (HCP, IXI) and \textit{in-vivo} (MR-ART) datasets. Qualitative comparisons against three state-of-the-art methods show that SSRL-MAR more effectively removes motion artifacts while preserving anatomical details.

In both simulated datasets (a-b), our method produces results closest to ground truth, with sharper cortical boundaries and reduced residual artifacts compared to alternatives. For \textit{in-vivo} MR-ART data (c), SSRL-MAR successfully handles both mild (L1) and severe (L2) motion corruption, demonstrating robust performance across motion severities. Visual inspection reveals that comparative methods either under-correct (leaving residual artifacts shown by black arrows) or over-smooth anatomical structures (shown by red arrows), while our approach maintains better texture preservation and structural fidelity.

\begin{figure}[t!]
	\centering
	\includegraphics[width=0.48\textwidth, draft=false]{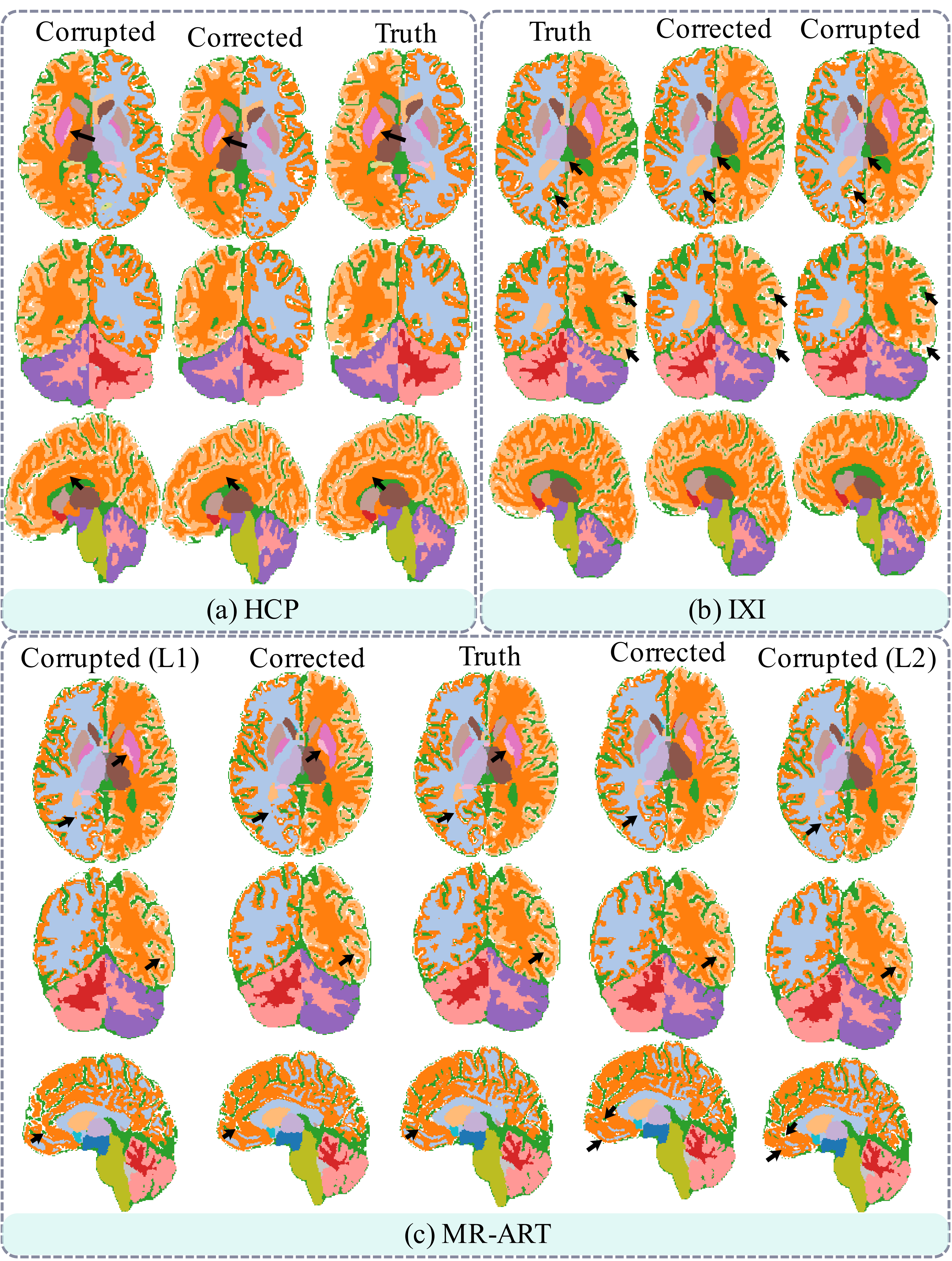}
	\caption{Volumetric segmentation consistency across orthogonal views. From left to right: truth (motion-free), corrupted input, and SSRL-MAR corrected output for (a) HCP, (b) IXI, and (c) MR-ART datasets. The corrected volumes maintain anatomical coherence across axial, sagittal, and coronal planes, demonstrating that SSRL-MAR leverages full 3D context rather than processing slices independently.}
	\label{fig:3d_consistency}
\end{figure}

To further illustrate the 3D capabilities of SSRL-MAR, Fig.~\ref{fig:3d_consistency} shows volumetric segmentation overlays across orthogonal views (axial, sagittal, coronal). Unlike 2D methods that process slices independently and introduce slice-wise artifacts (gold arrows in Fig.~\ref{fig:qualitative_fig01}), our method maintains anatomical coherence across all three planes. The corrected volumes preserve structural continuity from slice to slice, demonstrating that SSRL-MAR leverages full 3D context rather than treating the volume as independent 2D slices.

These consistent improvements across datasets and motion severities highlight the strong generalizability of the proposed framework for practical, real-world motion correction.

\subsubsection{Cortical Surface Analysis}\label{sec:results_surface}
 
\begin{figure}[t!] % or [b] or [!ht]
	\centering
	\includegraphics[width=.48\textwidth, draft=false]{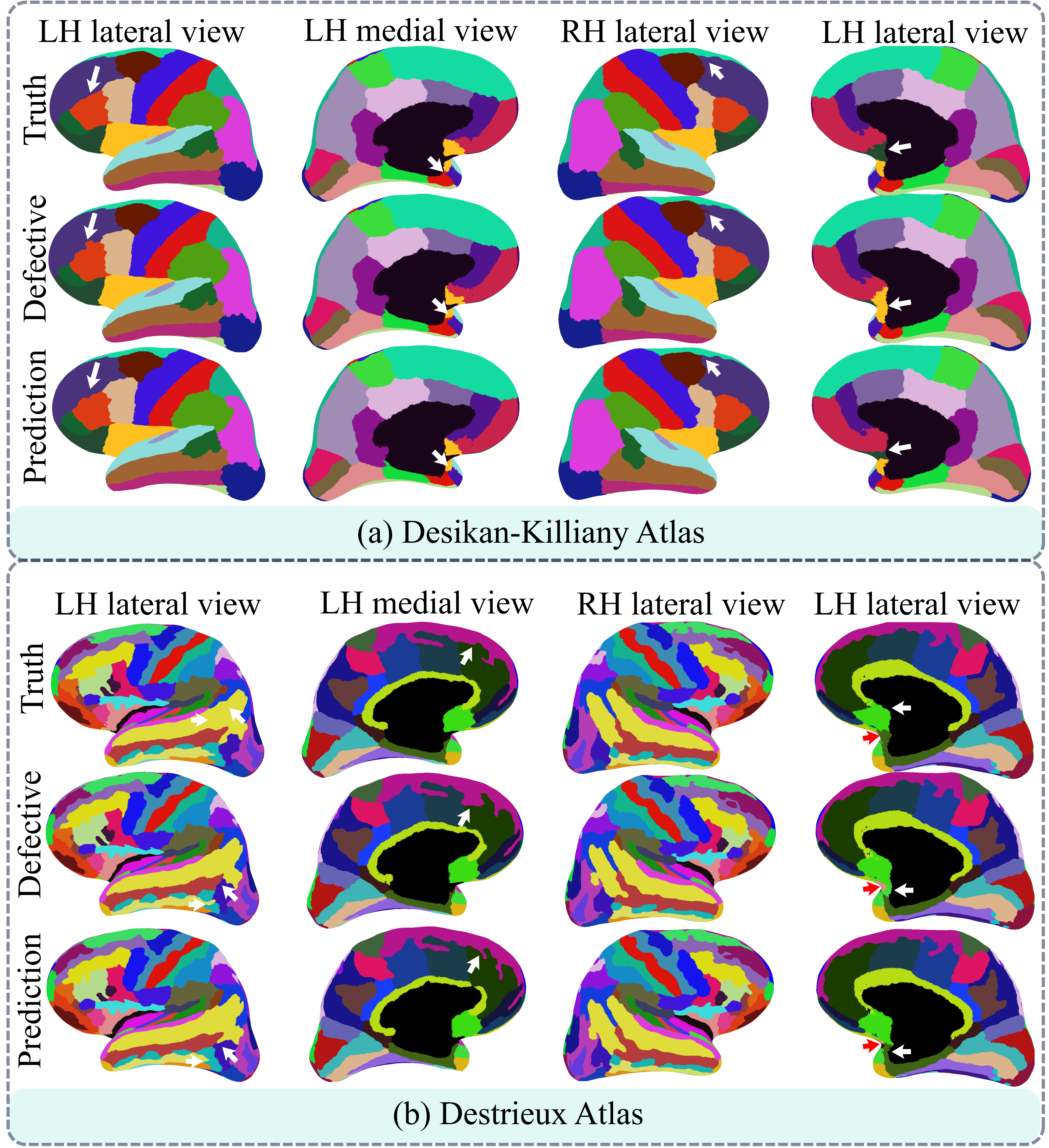}
	\caption{Cortical surface parcellations derived from SSRL-MAR-corrected reconstructions using the Desikan-Killiany (a) and Destrieux (b) atlases. Lateral and medial views of both hemispheres are shown. The corrected surfaces exhibit continuous gyri-sulci patterns and anatomically consistent regional boundaries. White arrows highlight motion-sensitive areas, particularly along the superior temporal, supramarginal, and parahippocampal cortices.}
	\label{fig:surface_cortex}
\end{figure}

Cortical surface parcellations derived from the motion-corrected images are presented in Fig.~\ref{fig:surface_cortex} for both the (a) Desikan-Killiany and (b) Destrieux atlas protocols. The resulting surfaces exhibit smooth and continuous gyral-sulcal boundaries without topological errors or fragmentation, confirming the preservation of anatomical fidelity. The reconstructions demonstrate accurate regional boundary delineation across both hemispheres in lateral and medial views, maintaining consistent anatomical correspondence.

Motion-sensitive regions, most prominently the superior temporal cortex, supramarginal gyrus, and parahippocampal cortex (white arrows), exhibit full recovery of sulcal depth, gyral curvature, and boundary continuity. These corrections restore the expected folding patterns that were distorted in the motion-corrupted inputs. The resulting surfaces demonstrate that SSRL-MAR maintains cortical topology essential for downstream morphometric tasks, including cortical thickness estimation, surface-based alignment, and atlas-based quantification.

\begin{table}[b!]
	\centering
	\caption{Quantitative evaluation of motion artifact reduction on \textit{in-silico} and \textit{in-vivo} datasets. Performance is measured for motion-corrupted (Uncorr.) and motion-corrected images across different motion levels. Two supervised variants are shown: oracle (fine-tuned on paired real data, impractical) and source-only (trained on simulated data only, practical baseline). Values are Mean $\pm$ SD.}
	\label{tab:image_metrics_qualitative}
	\resizebox{0.48\textwidth}{!}{%
		\begin{tabular}{llcccc}
			\hline
			\textbf{Metric} & Models & Mode &\textbf{in-silico} & \textbf{in-vivo (L1)} & \textbf{in-vivo (L2)} \\ \hline
			\multirow{6}{*}{PSNR (dB) $ \uparrow $} 
			&   Uncorr. & - & $21.87_{\pm 1.30}$ & $26.01_{\pm 2.74}$ & $24.50_{\pm2.72}$ \\
			&	CycleGAN  & 3D& $20.17_{\pm 1.12}$ & $25.50_{\pm 2.02}$ & $24.74_{\pm2.09}$ \\ 
			&	UDDN  & 2D& $20.90_{\pm 0.81}$ & $23.90_{\pm 3.16}$ & $23.76_{\pm3.82}$ \\ 
			&	AutoDPS & 2D & $20.07_{\pm 0.87}$ & $25.90_{\pm 2.87}$ & $25.11_{\pm 2.35}$ \\ 
			&	Supervised (oracle){$^\dagger$} & 3D & $24.56_{\pm 1.40 }$ & $27.37 _{\pm 2.39 }$ & $26.36 _{\pm 2.18 }$ \\
			&	Supervised (source-only){$^\ddagger$} & 3D & $24.56_{\pm 1.40 }$ & $25.05_{\pm 1.93 }$ & $24.09_{\pm 2.26 }$ \\ 
			&	Ours  & 3D& $23.81_{\pm1.45}$ & $26.90_{\pm 2.13}$ & $26.11_{\pm 2.24}$ \\ \hdashline
			\multirow{6}{*}{NMSE (\%) $ \downarrow $} 
			&   Uncorr. & - & $1.79_{\pm 0.33}$ & $0.63_{\pm 0.48}$ & $0.89_{\pm 0.65}$ \\
			&	CycleGAN & 3D & $1.22_{\pm 0.42}$ & $0.53_{\pm 0.38}$ & $0.81_{\pm 0.61}$ \\ 
			&	UDDN & 2D & $1.32_{\pm 0.36}$ & $0.73_{\pm0.58}$ & $0.90_{\pm 0.71}$ \\ 
			&	AutoDPS & 2D & $1.35_{\pm 0.41}$ & $0.50_{\pm 0.47}$ & $0.60_{\pm0.53}$ \\ 
			&	Supervised (oracle){$^\dagger$} & 3D & $0.66 _{\pm 0.21 }$ & $0.45 _{\pm 0.33 }$ & $0.54 _{\pm 0.35 }$ \\ 
			&	Supervised (source-only){$^\ddagger$} & 3D & $0.66 _{\pm 0.21 }$ & $0.65_{\pm 0.35 }$ & $0.85_{\pm 0.47 }$ \\ 
			&	Ours  & 3D& $0.79_{\pm0.25}$ & $0.48_{\pm0.33}$ & $0.58_{\pm 0.48}$ \\ \hdashline
			\multirow{6}{*}{SSIM (\%) $ \uparrow $} 
			&	Uncorr. & - & $88.54_{\pm 2.23}$ & $94.33_{\pm3.50}$ & $92.56_{\pm4.14}$ \\
			&	CycleGAN & 3D & $90.79_{\pm2.51}$ & $95.00_{\pm3.28}$ & $93.54_{\pm 3.77}$ \\ 
			&	UDDN & 2D & $89.07_{\pm 2.23}$ & $93.81_{\pm3.15}$ & $92.52_{\pm 3.61}$ \\ 
			&	AutoDPS & 2D & $89.51_{\pm 2.93}$ & $95.63_{\pm2.79}$ & $94.11_{\pm2.76}$ \\ 
			&	Supervised (oracle){$^\dagger$} & 3D & $92.58 _{\pm 1.75 }$ & $95.98 _{\pm 2.54 }$ & $95.06 _{\pm 2.70 }$ \\ 
			&	Supervised (source-only){$^\ddagger$} & 3D & $92.58 _{\pm 1.75 }$  & $93.96_{\pm 2.53  }$ & $92.75_{\pm 3.16 }$ \\ 
			&	Ours & 3D & $91.55_{\pm 2.01}$ & $95.82_{\pm2.50}$ & $95.03_{\pm 3.04}$ \\ \hline
		\end{tabular}%
	}
	\par\vspace{1mm}
	\begin{minipage}{0.48\textwidth}
		{\scriptsize
			{$^\dagger$}Pretrained on simulated \textit{in-silico} pairs, then fine-tuned on paired real MR-ART scans at each motion level; reported as an upper bound only, as such real pairs are unavailable in practice.\\
			{$^\ddagger$}Trained only on simulated \textit{in-silico} pairs and applied to in-vivo data without fine-tuning; the \textit{in-silico} column is therefore shared with the oracle base model.}
	\end{minipage}
\end{table}

\subsubsection{Quantitative Evaluation}\label{sec:results_quantitative}

The quantitative performance of the proposed SSRL-MAR framework is summarized in Table~\ref{tab:image_metrics_qualitative} for both \textit{in-silico} (HCP, IXI) and in-vivo (MR-ART) datasets. Across all datasets and motion levels, SSRL-MAR consistently enhanced image quality, achieving higher PSNR and SSIM values alongside lower NMSE compared to the motion-corrupted inputs.

On the \emph{in-silico} data, the method improved the average PSNR from 21.87 dB to 23.81 dB, reduced the NMSE from $ 1.79\% $ to $ 0.79\% $, and increased the SSIM from 88.54\% to 91.55\%. For the more challenging \textit{in-vivo} MR-ART cases, SSRL-MAR elevated the PSNR from 26.01 dB to 26.90 dB (L1) and from 24.50 dB to 26.11 dB (L2), with corresponding reductions in NMSE and consistent gains in SSIM. These results demonstrate that SSRL-MAR effectively suppresses motion artifacts while preserving structural fidelity and tissue contrast, yielding reconstructions that are quantitatively closer to the motion-free reference volumes.

To contextualize the performance of our self-supervised framework, we trained two supervised baselines using the same generator architecture. The first, ``Supervised (oracle)'' in Table~\ref{tab:image_metrics_qualitative}, was trained on \textit{in-silico} paired data and then fine-tuned on paired real MR-ART scans at each motion level using the same optimization settings as the generator training. This requires a separate fine-tuned model and paired real scans for each level, a scenario unavailable in practice, so we report it only as an upper bound. The second, ``Supervised (source-only)'', was trained once on \textit{in-silico} paired data and applied zero-shot to MR-ART without any target-domain fine-tuning, reflecting the realistic setting where only simulated paired data is available. Under this matched, zero-shot setting, SSRL-MAR before domain adaptation (Table~\ref{tab:ablation}) already exceeds the source-only supervised model by 0.54 dB (L1) and 0.63 dB (L2), indicating that direct supervision overfits the simulated artifact distribution and transfers poorly to real motion. Because SSRL-MAR is self-supervised, it can further adapt to the target domain using only unlabeled motion-corrupted scans; this raises its PSNR to 26.90 dB (L1) and 26.11 dB (L2), exceeding the source-only supervised model by 1.85 dB and 2.02 dB. The oracle supervised model remains higher (27.37/26.36 dB), but only because it is fine-tuned on paired real MR-ART scans that cannot be obtained in routine practice, and even then it leads our method by only 0.47 dB (L1) and 0.25 dB (L2). On \textit{in-silico} data the supervised models edge ours (24.56 vs. 23.81 dB) simply because both are trained and tested in the same simulated domain. Taken together, in every regime that is realizable in practice, SSRL-MAR matches or surpasses the supervised alternative while requiring no paired data.

\begin{table}[t!]
	\centering
	\caption{Segmentation-based relative volume error (\%) for selected structures before (Uncorr.) and after (Corr.) SSRL-MAR correction, compared against motion-free reference. Values are reported as Mean $\pm$ SD. CC = Corpus Callosum.}
	\label{tab:seg_relerr}
	\resizebox{0.48\textwidth}{!}{%
		\begin{tabular}{lccccc}
			\hline
			\textbf{Structure} & \multicolumn{2}{c}{\textbf{Level 1}} & & \multicolumn{2}{c}{\textbf{Level 2}} \\
			\cline{2-3}\cline{5-6}
			& \textbf{Uncorr.} & \textbf{Corr.} & &\textbf{Uncorr.} & \textbf{Corr.} \\ \hline
			CC Central        & $6.22_{\pm 15.43}$ & $2.83_{\pm 14.01}$ && $4.83_{\pm 28.98}$ & $2.95_{\pm 18.71}$ \\
			3rd-Ventricle      & $4.23_{\pm 6.33}$   & $1.21_{\pm 5.18}$ & & $1.98_{\pm 7.22}$  & $1.63_{\pm 4.70}$  \\
			4th-Ventricle      & $2.40_{\pm 2.26}$   & $1.15_{\pm 2.24}$ &  & $0.31_{\pm 5.35}$  & $-0.96_{\pm 5.92}$ \\
			Brain-Stem         & $0.66_{\pm 1.17}$   & $-0.22_{\pm 1.02}$& & $0.35_{\pm 1.10}$  & $-0.34_{\pm 0.97}$ \\
			CC Mid Anterior  & $1.16_{\pm 13.92}$ & $0.93_{\pm 12.51}$& & $-4.33_{\pm 4.01}$ & $-3.98_{\pm 3.38}$ \\
			CC Anterior       & $1.17_{\pm 5.88}$  & $1.21_{\pm 6.17}$ & & $1.72_{\pm 9.49}$  & $0.88_{\pm 7.54}$  \\ \hline
		\end{tabular}%
	}
\end{table}

\begin{figure}[b!] 
	\centering
	\includegraphics[width=0.48\textwidth]{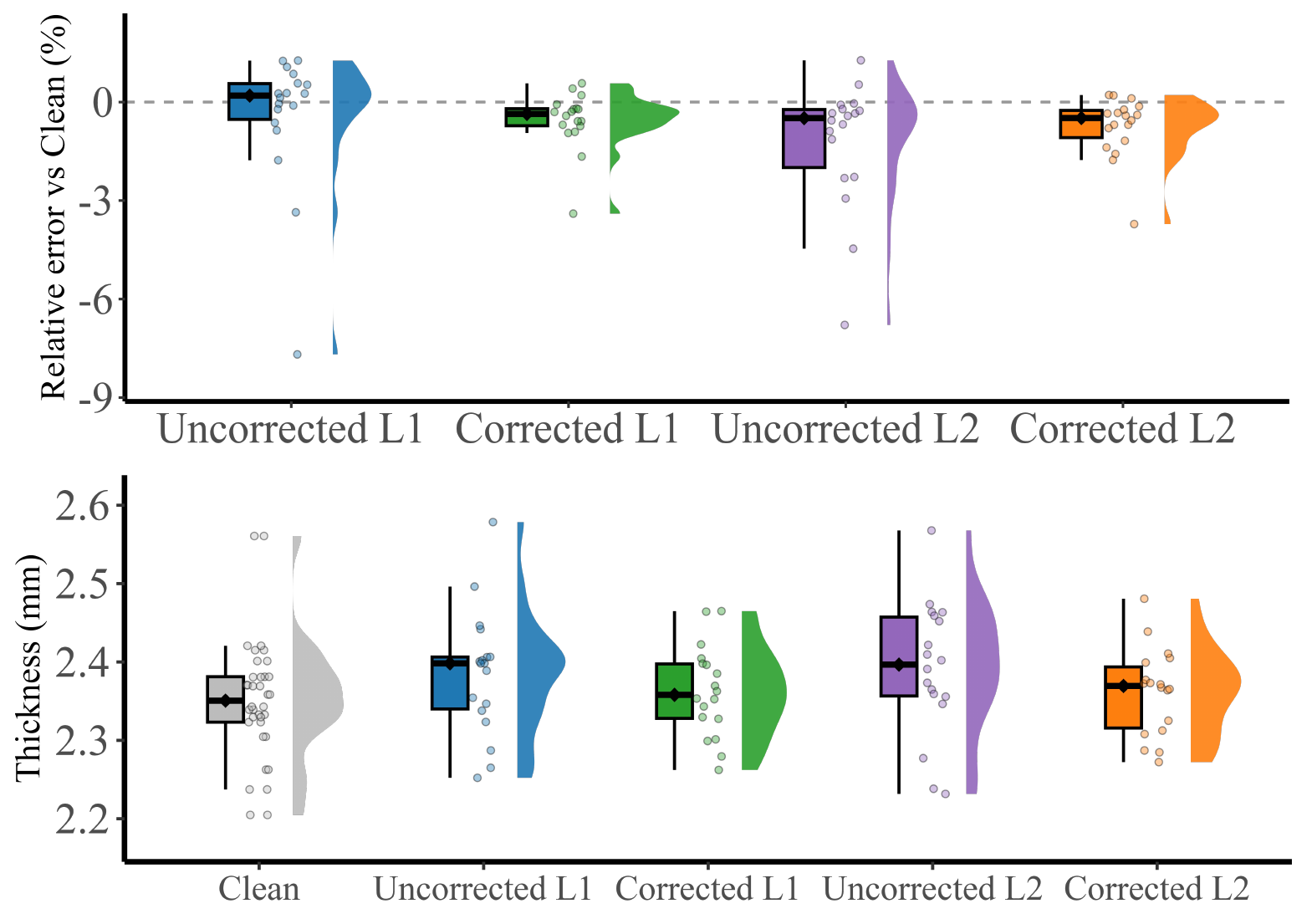}   
	\caption{Distribution of motion-induced volumetric errors across cortical regions. Violin and box plots illustrate the relative error for uncorrected (Uncorr.) and SSRL-MAR corrected images at L1 and L2 motion levels. (a) Error reduction relative to the motion-corrupted state. (b) Final accuracy of corrected outputs versus the motion-free reference.}
	\label{fig:cortexThickness_segDiff}
\end{figure}

\subsubsection{Robustness Across Motion Regimes}\label{sec:robustnessExtremeMotionSeverity}

We evaluated SSRL-MAR under two complementary protocols to characterize its behavior across motion severities. In the matched protocol, which probes per-regime capacity, the full three-stage pipeline was retrained from scratch on \textit{in-silico} data corrupted at a given range and then tested at that same range, using the identical 10-event protocol and training hyperparameters described in Section~\ref{sec:progressive_training_strategy}. In the zero-shot protocol, the model trained at the $\pm2$ range was applied directly to data corrupted at larger ranges, reflecting the deployment setting in which a single fixed model encounters motion more severe than that seen during training. Each range jointly scales rotation in degrees and translation in millimeters.

\begin{table}[!t]
	\centering
	\caption{Robustness of SSRL-MAR to motion severity under two regimes: \emph{zero-shot}, where the model trained at the $\pm2$ range is applied to larger ranges, and \emph{matched}, where the model is trained and tested at the same range. Results are reported as mean $\pm$ SD on the \textit{in-silico} validation set using the same 10-event protocol as in training. Each range jointly scales rotation (degrees) and translation (mm).}
	\label{tab:robustness}
	\resizebox{0.48\textwidth}{!}{%
		\begin{tabular}{llccc}
			\hline
			Regime & Range & PSNR (dB) $\uparrow$ & SSIM (\%) $\uparrow$ & NMSE (\%) $\downarrow$ \\ \hline
			\multirow{6}{*}{Zero-shot ($\pm2$){$^\dagger$}}
			& $\pm2$ & $23.81_{\pm1.44}$ & $91.55_{\pm2.00}$ & $0.79_{\pm0.25}$ \\
			& $\pm3$ & $21.76_{\pm1.61}$ & $87.50_{\pm3.18}$ & $1.29_{\pm0.51}$ \\
			& $\pm4$ & $21.14_{\pm1.62}$ & $86.09_{\pm3.44}$ & $1.49_{\pm0.55}$ \\
			& $\pm5$ & $20.96_{\pm1.68}$ & $85.74_{\pm3.65}$ & $1.57_{\pm0.66}$ \\
			& $\pm6$ & $20.61_{\pm1.68}$ & $84.95_{\pm3.93}$ & $1.69_{\pm0.66}$ \\
			& $\pm7$ & $20.53_{\pm1.82}$ & $84.70_{\pm4.15}$ & $1.74_{\pm0.70}$ \\ \hdashline
			\multirow{3}{*}{Matched{$^\ddagger$}}
			& $\pm3$ & $22.94_{\pm1.65}$ & $88.21_{\pm3.12}$ & $1.02_{\pm0.38}$ \\
			& $\pm5$ & $21.78_{\pm1.89}$ & $85.93_{\pm3.78}$ & $1.35_{\pm0.52}$ \\
			& $\pm7$ & $21.02_{\pm2.14}$ & $84.12_{\pm4.01}$ & $1.58_{\pm0.61}$ \\ \hline
		\end{tabular}%
	}
	\par\vspace{1mm}
	\begin{minipage}{0.48\textwidth}
		{\scriptsize {$^\dagger$}The matched $\pm2$ result coincides with the zero-shot $\pm2$ entry, as both use the model trained at $\pm2$; it is listed once under the zero-shot block.\\{$^\ddagger$}Matched models for $\pm3$, $\pm5$, and $\pm7$ were retrained from scratch with the same three-stage pipeline and hyperparameters as in Section~\ref{sec:progressive_training_strategy}.}
	\end{minipage}
\end{table}

Under matched training and testing (Table~\ref{tab:robustness}), SSRL-MAR achieved 22.94 dB at $\pm3$, 21.78 dB at $\pm5$, and 21.02 dB at $\pm7$. At every severity these values exceed the corresponding zero-shot results in PSNR, by 1.18 dB ($\pm3$), 0.82 dB ($\pm5$), and 0.49 dB ($\pm7$), with consistently lower NMSE (1.02 vs. 1.29, 1.35 vs. 1.57, and 1.58 vs. 1.74). Because the motion-corrupted input at each range is identical across the two protocols, the higher matched PSNR corresponds to a larger relative improvement over the corrupted input at every severity. These results indicate that the architecture has sufficient capacity to correct severe motion when that regime is represented during training, and that the progressive decline observed under the zero-shot protocol was driven primarily by the train-test distribution shift rather than by an intrinsic ceiling of the model.

The benefit of matched training was largest at moderate severity and compressed at the most extreme range. In SSIM, matched training improved over zero-shot at $\pm3$ (88.21 vs. 87.50) and $\pm5$ (85.93 vs. 85.74), whereas at $\pm7$ SSIM was marginally lower (84.12 vs. 84.70), a difference of about 0.6 percentage points that is small relative to the roughly 4\% standard deviation of both measurements. This pattern is consistent with the behavior of the two metrics near the recoverable limit: at $\pm7$ the motion destroys high-frequency structure that no training distribution can restore, so structural similarity saturates, while PSNR and NMSE, which track voxelwise error, continue to favor matched training. The $\pm7$ crossover therefore does not indicate a failure of matched training; it marks the severity at which the recoverable signal, rather than the training distribution, becomes the binding constraint.

Taken together, the two protocols show that SSRL-MAR does not exhibit a sharp breaking point across the tested range. The matched analysis confirms that severe motion is correctable when represented in training, and the zero-shot analysis confirms that a single deployed model degrades gracefully and continues to deliver meaningful artifact reduction under motion well beyond its training range. This establishes the method's applicability across a wide spectrum of motion severities, in both retrainable and fixed-model deployment scenarios.

\subsubsection{Quantitative Segmentation Analysis}

Quantitative segmentation analysis confirmed that SSRL-MAR substantially reduces motion-induced volumetric bias, particularly in anatomically vulnerable regions. As quantified in Table~\ref{tab:seg_relerr}, the most pronounced error reductions were observed in the central corpus callosum and ventricular structures at the milder motion level, where the mean relative error decreased by more than 50\% following correction. At the more severe motion level, improvements in these same structures were smaller and less consistent, particularly in the 4th-Ventricle. In contrast, improvements in global volumes and the brainstem were more modest, which is consistent with their lower inherent susceptibility to motion-induced blurring and the known variability of automated segmentation pipelines. These results collectively demonstrate that SSRL-MAR mitigates motion-related distortions effectively in critical brain structures while maintaining volumetric consistency across the entire brain.

The distribution of regional segmentation errors is further detailed in Fig.~\ref{fig:cortexThickness_segDiff}. The violin and box plots illustrate that motion corruption (Uncorrected, Levels 1 and 2) introduces a systematic underestimation of cortical volumes. SSRL-MAR correction substantially mitigates this bias, shifting the error distribution centrally toward zero. Furthermore, the corrected outputs exhibit a tighter dispersion and reduced bias when compared directly to the clean references, confirming the method's efficacy in restoring faithful cortical morphometry across a range of motion severities.

\begin{figure}[b!] 
	\centering
	\includegraphics[width=0.48\textwidth, draft=false]{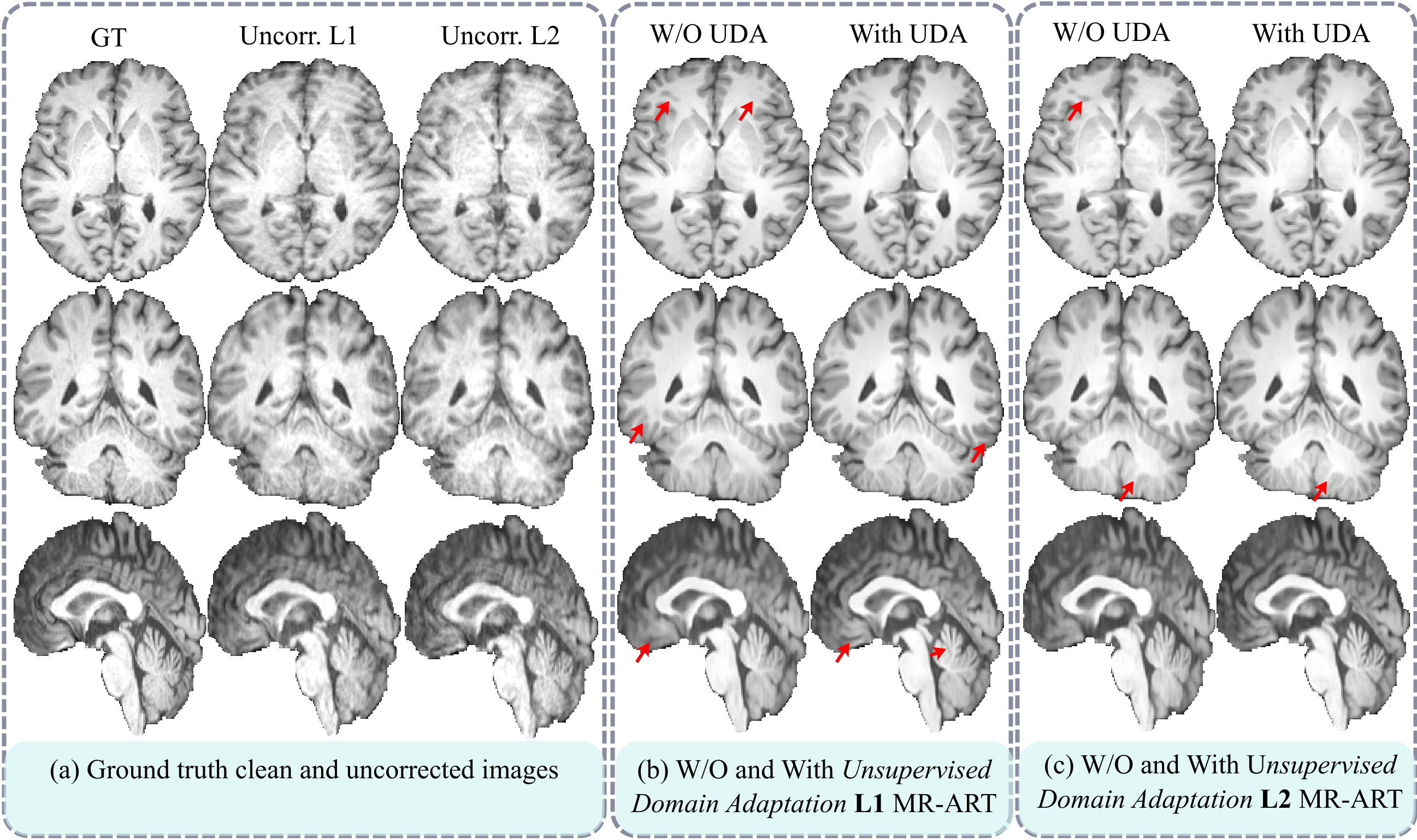}
    \caption{Effect of generalization and unsupervised domain adaptation on \textit{in-vivo} MR-ART reconstructions. The model trained solely on \textit{in-silico} data already suppresses motion-induced blur and ghosting, while unsupervised domain adaptation further sharpens cortical boundaries and reduces residual distortions (red arrows) in motion-sensitive regions.}
	\label{fig:ablation_mrART_fine_tune}
\end{figure}

\subsection{Ablation Study}\label{sec:ablation_study}

A series of ablation experiments were conducted to assess the contribution of each core component of the SSRL-MAR framework and to evaluate the impact of unsupervised domain adaptation on the \textit{in-vivo} MR-ART dataset. Quantitative results are summarized in Table~\ref{tab:ablation}.

\subsubsection{Unsupervised Domain Adaptation on \textit{in-vivo} MR-ART Data}

The generalization capability of SSRL-MAR was evaluated by applying the model, pretrained on \textit{in-silico} data, directly to the \textit{in-vivo} MR-ART dataset. Without any adaptation, the model effectively reduced motion-induced artifacts such as blurring and ghosting, producing reconstructions with improved structural clarity compared to the uncorrected inputs (Fig.~\ref{fig:ablation_mrART_fine_tune}). Quantitatively, this base model achieved PSNR values of $25.59$ dB (L1) and $24.72$ dB (L2), demonstrating robust zero-shot generalization to real-world motion patterns.

To investigate whether performance could be further improved by adapting to the target domain, we performed unsupervised domain adaptation using only the motion-corrupted volumes from the MR-ART training split. No motion-free references were used during this process. The contrastive encoder was further trained for 50 epochs on real \textit{in-vivo} corrupted patches to adapt the motion representation to the target artifact distribution, using the same optimizer, learning rate, and contrastive loss as in Stage 1. The degrader and generator were then updated for 50 epochs with the identical hyperparameters and loss functions employed in Stages 2 and 3, respectively, while the adapted encoder provided motion representations conditioned on real artifacts. This entire adaptation procedure required no paired data and preserved the fully unsupervised nature of the framework.

After adaptation, residual artifacts were further suppressed, resulting in enhanced delineation of cortical boundaries, particularly in motion-sensitive regions such as the ventricles and temporal lobes (Fig.~\ref{fig:ablation_mrART_fine_tune}). The adapted model yielded improved metrics, with PSNR increasing to $26.90$ dB (L1) and $26.11$ dB (L2). These gains confirm that unsupervised domain adaptation enhances the recovery of local anatomical detail and overall fidelity when moving from simulated to real motion artifacts, while maintaining the core unpaired paradigm of the framework.

\subsubsection{Component-wise Ablation on \textit{In-silico} Data}

Controlled experiments on the \textit{in-silico} dataset were conducted to quantify the contribution of each core component. The most substantial performance degradation occurred upon removing the CL (w/o CL.) module, where replacing the learned motion representation with Gaussian noise caused the PSNR to drop to $ 20.40 $ dB. This result confirms that the motion-sensitive latent space provides essential features that cannot be substituted by unstructured noise.

Ablating the MAA (w/o MAA) blocks, by substituting them with standard convolutional layers in the degrader network, led to a marked decline in performance, reducing the PSNR to $ 21.79 $ dB. This underscores the critical role of the MAA mechanism in generating realistic, structured motion artifacts necessary for effective generator training.

The removal of both the adversarial and gradient-consistency losses (w/o GAN + Grad) yielded overly smooth reconstructions, with a PSNR of $ 22.52 $ dB, confirming their combined role in preserving perceptual realism and anatomical sharpness. Isolating the removal of the gradient-consistency loss (w/o Grad) further resulted in a measurable decline to $ 23.01 $ dB, highlighting its specific contribution to stabilizing the recovery of fine structural boundaries.

Ultimately, the full SSRL-MAR model achieved optimal performance, with a PSNR of $ 23.81 $ dB and an SSIM of $ 91.55\% $. The progressive improvement observed from each ablated variant to the complete model demonstrates that each component addresses a distinct aspect of the motion correction problem, culminating in the most effective artifact suppression and structural preservation.

To further isolate the contribution of the anisotropic loss component, we compared our full anisotropic InfoNCE loss against a standard isotropic InfoNCE loss. The isotropic variant achieved 23.42 dB PSNR, 90.78\% SSIM, and 0.87\% NMSE, while the anisotropic loss improved results to 23.81 dB PSNR, 91.55\% SSIM, and 0.79\% NMSE. This confirms that dimension reweighting enhances motion representation learning by emphasizing feature directions most discriminative for motion versus clean content. These results are included in Table~\ref{tab:ablation} for completeness.

\begin{table}[t!]
	\centering
	\caption{Ablation results of the proposed SSRL-MAR model on the \textit{in-silico} and \textit{in-vivo} MR-ART datasets. Values represent the mean $\pm$ SD.}
	
	\label{tab:ablation}
	\centering
	\resizebox{0.48\textwidth}{!}{%
		\begin{tabular}{lccc}
			\hline
			\textbf{Configuration} & \textbf{PSNR (dB) $ \uparrow $} & \textbf{NMSE (\%) $ \downarrow $} & \textbf{SSIM (\%)$ \uparrow $} \\
			\hline
			\multicolumn{4}{l}{\textbf{\textit{In-silico} (component-wise ablation)}} \\
			w/o CL & $20.40_{\pm {2.21}}$ & $1.20_{\pm {0.58}}$ & $87.80_{\pm {3.47}}$ \\
			w/o MAA & $21.79_{\pm {1.27}}$ & $1.24_{\pm {0.34}}$ & $89.91_{\pm {2.27}}$ \\
			w/o GAN\,+\,Grad & $22.52_{\pm {1.12}}$ & $1.03_{\pm {0.32}}$ & $90.29_{\pm {2.12}}$ \\
			w/o Grad & $23.01_{\pm {1.20}}$ & $0.86_{\pm {0.32}}$ & $90.89_{\pm {2.18}}$ \\
			Standard InfoNCE & $23.42_{\pm 1.38}$ & $0.87_{\pm 0.28}$ & $90.78_{\pm 2.11}$ \\   %%%%%%%% TODO
			\textbf{Full SSRL-MAR} & ${23.81}_{\pm 1.45}$ & ${0.79}_{\pm 0.25}$ & ${91.55}_{\pm 2.01}$ \\
			\hdashline
			\multicolumn{4}{l}{\textbf{\textit{In-vivo} MR-ART (unsupervised domain adaptation (UDA) ablation)}} \\
			Uncorr. L1 & $26.01_{\pm 2.74}$ & $0.63_{\pm 0.48}$ & $94.33_{\pm 3.50}$ \\
			Corr. L1 (w/o UDA) & $ 25.59_{\pm {1.71}} $ & $0.62_{\pm {0.34}} $ & $94.41_{\pm {2.50}} $ \\
			Corr. L1 (w UDA) & ${26.90}_{\pm 2.13}$ & ${0.48}_{\pm 0.33}$ & ${95.82}_{\pm 2.50}$ \\
			Uncorr. L2 & $24.50_{\pm 2.72}$ & $0.89_{\pm 0.65}$ & $92.56_{\pm 4.14}$ \\
			Corr. L2 (w/o UDA) & $24.72_{\pm {2.01}} $ & $0.78_{\pm 0.47} $ & $93.31_{\pm {3.13}} $ \\
			Corr. L2 (w UDA) & ${26.11}_{\pm 2.24}$ & ${0.58}_{\pm 0.48}$ & ${95.03}_{\pm 3.04}$ \\
			\hline
		\end{tabular}%
	}
\end{table}

\section{Discussion and Conclusion}\label{sec:discussion_conclusion}

The proposed SSRL-MAR framework provides an effective solution for reducing motion artifacts in 3D brain MRI without requiring paired supervision or raw k-space data. Across simulated and \textit{in-vivo} datasets, the method consistently improves image quality, preserves anatomical detail, and maintains cortical topology in downstream surface reconstructions. These results highlight the advantages of combining contrastive representation learning, degradation-aware simulation, and restoration within a unified self-supervised framework.

The ablation experiments clarify the contribution of each component. The contrastive pretraining stage proves essential for learning motion-sensitive representations that separate anatomical content from motion-induced distortions. The degrader network, guided by these latent features, generates realistic artifact patterns that approximate real motion distributions, enabling reliable self-supervision for the generator. The adversarial and gradient-consistency terms further enhance structural sharpness and edge integrity, preventing the over-smoothing observed when these losses are removed. Together, these elements form a stable and robust training pipeline capable of modeling diverse motion patterns encountered across sites.

A natural benchmark for our method is supervised training on paired data, and we considered two regimes that bracket what is achievable in practice. The source-only supervised model, trained on the same simulated pairs as SSRL-MAR, defines the realizable setting in which clean references exist only in simulation. Under this matched regime, direct supervision overfits the simulated artifact distribution and transfers poorly to real motion, and SSRL-MAR generalizes more effectively to \textit{in-vivo} MR-ART data even before any target adaptation. Because the self-supervised formulation does not depend on clean references, it can additionally exploit unlabeled motion-corrupted real volumes through unsupervised domain adaptation, a route unavailable to supervised training, which further widens this margin. The oracle supervised model, fine-tuned on paired real MR-ART scans, retains a small advantage of 0.25 to 0.47 dB PSNR, but this gap is attainable only with simultaneously acquired motion-free and motion-corrupted scans of the same subject, which are absent from retrospective archives and routine clinical workflows. Across every regime that can be realized in practice, SSRL-MAR therefore matches or exceeds supervised correction while requiring no paired data, which is the property that determines applicability at population scale.

A critical consideration for clinical adoption is whether the model might introduce hallucinated structures. Several aspects of SSRL-MAR mitigate this risk. The anatomy-invariant contrastive encoder prevents the generator from accessing subject-specific anatomical information that could be used to invent false features. The gradient-consistency loss anchors reconstructions to observed edges, preserving authentic sulcal and gyral boundaries. The ablation study confirms that without these constraints, the model defaults to over-smoothing (PSNR 22.52 dB in the w/o GAN + Grad condition) rather than hallucination, suggesting that the framework prioritizes conservative restoration when uncertain. Furthermore, FreeSurfer-based cortical analysis revealed continuous, topologically correct surfaces (Fig.~\ref{fig:surface_cortex}) without the fragmentation or implausible folds that would indicate hallucination. The quantitative segmentation analysis (Table~\ref{tab:seg_relerr}) further demonstrates that volumetric errors are systematically reduced, not randomly redistributed. The 3D consistency of the reconstructions is further illustrated in Fig.~\ref{fig:3d_consistency}, which shows coherent anatomical structures across orthogonal views, confirming that SSRL-MAR leverages full volumetric context rather than processing slices independently. The comparison with 2D methods adapted via slice-wise processing highlights a key advantage of native 3D approaches like SSRL-MAR. As shown in Fig.~\ref{fig:qualitative_fig01}, 2D methods introduce slice-wise artifacts (gold arrows) because they cannot enforce volumetric consistency. This dimensional mismatch underscores the importance of full 3D processing for anatomically coherent motion correction. These multiple safeguards and validation steps support the anatomical fidelity of SSRL-MAR outputs.

Generalization to \textit{in-vivo} MR-ART data demonstrates the practical utility of SSRL-MAR. Even without target-domain adaptation, the model trained purely on simulated data reduces blur and ghosting in real motion-corrupted scans, indicating that the learned motion representations capture domain-invariant characteristics. Unsupervised domain adaptation yields additional improvements in boundary definition and local detail, particularly in motion-sensitive cortical regions. Critically, no paired data is used at any stage of this process: pretraining relies on unpaired clean and motion-corrupted volumes, and domain adaptation uses only unlabeled, motion-corrupted scans from the MR-ART training split, with no motion-free counterpart. This is precisely the setting in which paired supervision cannot be obtained, since simultaneously acquired motion-free and motion-corrupted scans of the same subject are unavailable outside dedicated repeat-acquisition studies. The FreeSurfer-based volumetric and topological analyses described above were obtained under this same unpaired regime, indicating that the anatomical fidelity of the correction does not depend on access to paired real training data.

Although the framework performs well across a range of conditions, several limitations warrant consideration. The study focuses on T1w imaging; extending the approach to additional contrasts such as T2-weighted or FLAIR, as well as to non-brain anatomies, would require further validation and potentially contrast-adaptive modeling. While our synthetic motion generation using TorchIO produces artifacts that visually and quantitatively approximate real motion patterns, a synthetic-to-real gap still exists. The direct generalization of the pretrained model to the MR-ART dataset suggests that the learned representation captures domain-invariant aspects of motion corruption, but incorporating physics-informed priors or sparse k-space information may further improve realism and cross-domain robustness. In addition, we evaluated robustness under both matched and zero-shot motion regimes. Matched training confirmed that severe motion is correctable when represented during training, with PSNR and NMSE improving over the zero-shot model at every severity, while absolute quality still decreased as motion increased under both regimes. This indicates that the residual decline at high severity reflects the diminishing recoverable signal rather than distribution shift alone, and that SSRL-MAR degrades gracefully without an abrupt failure point. Broader modeling of extreme and complex intra-volume motion, particularly highly non-rigid patterns, together with training pools spanning a wider range of severities, could further improve robustness.

In conclusion, SSRL-MAR provides a scalable and anatomically reliable framework for 3D brain MRI motion artifact reduction using fully unpaired image-domain supervision. By unifying motion representation learning, artifact synthesis, and motion-aware restoration, the method achieves robust correction on both simulated and real datasets while preserving anatomically meaningful detail. These results support its use in retrospective neuroimaging pipelines aimed at improving downstream morphometric and surface-based analyses in large-scale clinical and research cohorts.

\bibliography{ssrl_moco.bib}

@article{https://doi.org/10.1002/jmri.24850,
	author = {Zaitsev, Maxim and Maclaren, Julian and Herbst, Michael},
	title = {Motion artifacts in MRI: A complex problem with many partial solutions},
	journal = {Journal of Magnetic Resonance Imaging},
	volume = {42},
	number = {4},
	pages = {887-901},
	doi = {https://doi.org/10.1002/jmri.24850},
	url = {https://onlinelibrary.wiley.com/doi/abs/10.1002/jmri.24850},
	eprint = {https://onlinelibrary.wiley.com/doi/pdf/10.1002/jmri.24850},
	year = {2015}
}

@misc{safari2025systematicreviewmetaanalysisaidriven,
	title = {Systematic review and meta-analysis of AI-driven MRI motion artifact detection and correction},
	journal = {Physica Medica},
	volume = {141},
	pages = {105704},
	year = {2026},
	issn = {1120-1797},
	doi = {https://doi.org/10.1016/j.ejmp.2025.105704},
	url = {https://www.sciencedirect.com/science/article/pii/S1120179725008142},
	author = {Mojtaba Safari and Zach Eidex and Richard L.J. Qiu and Matthew Goette and Tonghe Wang and Xiaofeng Yang},
}

@article{fu2020deep,
	title={Deep learning in medical image registration: a review},
	author={Fu, Yabo and Lei, Yang and Wang, Tonghe and Curran, Walter J and Liu, Tian and Yang, Xiaofeng},
	journal={Physics in Medicine and Biology},
	volume={65},
	number={20},
	pages={20TR01},
	year={2020},
	publisher={IOP Publishing}
}

@article{LIN2006751,
	title = {Improved optimization strategies for autofocusing motion compensation in MRI via the analysis of image metric maps},
	journal = {Magnetic Resonance Imaging},
	volume = {24},
	number = {6},
	pages = {751-760},
	year = {2006},
	issn = {0730-725X},
	doi = {https://doi.org/10.1016/j.mri.2006.02.003},
	url = {https://www.sciencedirect.com/science/article/pii/S0730725X06001445},
	author = {Wei Lin and Hee Kwon Song}
}

@article{https://doi.org/10.1002/mrm.24463,
	author = {Usman, Muhammad and Atkinson, David and Odille, Freddy and Kolbitsch, Christoph and Vaillant, Ghislain and Schaeffter, Tobias and Batchelor, Philip G. and Prieto, Claudia},
	title = {Motion corrected compressed sensing for free-breathing dynamic cardiac MRI},
	journal = {Magnetic Resonance in Medicine},
	volume = {70},
	number = {2},
	pages = {504-516},
	doi = {https://doi.org/10.1002/mrm.24463},
	url = {https://onlinelibrary.wiley.com/doi/abs/10.1002/mrm.24463},
	eprint = {https://onlinelibrary.wiley.com/doi/pdf/10.1002/mrm.24463},
	year = {2013}
}

@ARTICLE{8252880,
	author={Haskell, Melissa W. and Cauley, Stephen F. and Wald, Lawrence L.},
	journal={IEEE Transactions on Medical Imaging}, 
	title={TArgeted Motion Estimation and Reduction (TAMER): Data Consistency Based Motion Mitigation for MRI Using a Reduced Model Joint Optimization}, 
	year={2018},
	volume={37},
	number={5},
	pages={1253-1265},
	doi={10.1109/TMI.2018.2791482}}

@ARTICLE{10285512,
	author={Spieker, Veronika and Eichhorn, Hannah and Hammernik, Kerstin and Rueckert, Daniel and Preibisch, Christine and Karampinos, Dimitrios C. and Schnabel, Julia A.},
	journal={IEEE Transactions on Medical Imaging}, 
	title={Deep Learning for Retrospective Motion Correction in MRI: A Comprehensive Review}, 
	year={2024},
	volume={43},
	number={2},
	pages={846-859},
	doi={10.1109/TMI.2023.3323215}}

@article{DUFFY2021117756,
	title = {Retrospective motion artifact correction of structural MRI images using deep learning improves the quality of cortical surface reconstructions},
	journal = {NeuroImage},
	volume = {230},
	pages = {117756},
	year = {2021},
	issn = {1053-8119},
	doi = {https://doi.org/10.1016/j.neuroimage.2021.117756},
	url = {https://www.sciencedirect.com/science/article/pii/S1053811921000331},
	author = {Ben A Duffy and Lu Zhao and Farshid Sepehrband and Joyce Min and Danny JJ Wang and Yonggang Shi and Arthur W Toga and Hosung Kim}
}

@article{https://doi.org/10.1002/mrm.28991,
	author = {Slipsager, Jakob M. and Glimberg, Stefan L. and Højgaard, Liselotte and Paulsen, Rasmus R. and Wighton, Paul and Tisdall, M. Dylan and Jaimes, Camilo and Gagoski, Borjan A. and Grant, P. Ellen and van der Kouwe, André and Olesen, Oline V. and Frost, Robert},
	title = {Comparison of prospective and retrospective motion correction in 3D-encoded neuroanatomical MRI},
	journal = {Magnetic Resonance in Medicine},
	volume = {87},
	number = {2},
	pages = {629-645},
	doi = {https://doi.org/10.1002/mrm.28991},
	url = {https://onlinelibrary.wiley.com/doi/abs/10.1002/mrm.28991},
	eprint = {https://onlinelibrary.wiley.com/doi/pdf/10.1002/mrm.28991},
	year = {2022}
}

@article{https://doi.org/10.1002/mrm.29255,
	author = {Brackenier, Yannick and Cordero-Grande, Lucilio and Tomi-Tricot, Raphael and Wilkinson, Thomas and Bridgen, Philippa and Price, Anthony and Malik, Shaihan J. and De Vita, Enrico and Hajnal, Joseph V.},
	title = {Data-driven motion-corrected brain MRI incorporating pose-dependent B0 fields},
	journal = {Magnetic Resonance in Medicine},
	volume = {88},
	number = {2},
	pages = {817-831},
	doi = {https://doi.org/10.1002/mrm.29255},
	url = {https://onlinelibrary.wiley.com/doi/abs/10.1002/mrm.29255},
	eprint = {https://onlinelibrary.wiley.com/doi/pdf/10.1002/mrm.29255},
	year = {2022}
}

@article{https://doi.org/10.1002/mrm.27783,
	author = {Küstner, Thomas and Armanious, Karim and Yang, Jiahuan and Yang, Bin and Schick, Fritz and Gatidis, Sergios},
	title = {Retrospective correction of motion-affected MR images using deep learning frameworks},
	journal = {Magnetic Resonance in Medicine},
	volume = {82},
	number = {4},
	pages = {1527-1540},
	doi = {https://doi.org/10.1002/mrm.27783},
	url = {https://onlinelibrary.wiley.com/doi/abs/10.1002/mrm.27783},
	eprint = {https://onlinelibrary.wiley.com/doi/pdf/10.1002/mrm.27783},
	year = {2019}
}

@article{https://doi.org/10.1002/mrm.29188,
	author = {Xu, Xiaojian and Kothapalli, Satya V. V. N. and Liu, Jiaming and Kahali, Sayan and Gan, Weijie and Yablonskiy, Dmitriy A. and Kamilov, Ulugbek S.},
	title = {Learning-based motion artifact removal networks for quantitative R2s mapping},
	journal = {Magnetic Resonance in Medicine},
	volume = {88},
	number = {1},
	pages = {106-119},
	doi = {https://doi.org/10.1002/mrm.29188},
	url = {https://onlinelibrary.wiley.com/doi/abs/10.1002/mrm.29188},
	eprint = {https://onlinelibrary.wiley.com/doi/pdf/10.1002/mrm.29188},
	year = {2022}
}

@article{https://doi.org/10.1002/mp.16844,
	author = {Safari, Mojtaba and Yang, Xiaofeng and Fatemi, Ali and Archambault, Louis},
	title = {MRI motion artifact reduction using a conditional diffusion probabilistic model (MAR-CDPM)},
	journal = {Medical Physics},
	volume = {51},
	number = {4},
	pages = {2598-2610},
	doi = {https://doi.org/10.1002/mp.16844},
	url = {https://aapm.onlinelibrary.wiley.com/doi/abs/10.1002/mp.16844},
	eprint = {https://aapm.onlinelibrary.wiley.com/doi/pdf/10.1002/mp.16844},
	year = {2024}
}

@Article{Liu2021_nature_MI,
	author={Liu, Siyuan
	and Thung, Kim-Han
	and Qu, Liangqiong
	and Lin, Weili
	and Shen, Dinggang
	and Yap, Pew-Thian},
	title={Learning MRI artefact removal with unpaired data},
	journal={Nature Machine Intelligence},
	year={2021},
	month={Jan},
	day={01},
	volume={3},
	number={1},
	pages={60-67},
	issn={2522-5839},
	doi={10.1038/s42256-020-00270-2},
	url={https://doi.org/10.1038/s42256-020-00270-2}
}

@article{Safari_2024_maudgan,
	doi = {10.1088/1361-6560/ad4845},
	url = {https://doi.org/10.1088/1361-6560/ad4845},
	year = {2024},
	month = {may},
	publisher = {IOP Publishing},
	volume = {69},
	number = {11},
	pages = {115057},
	author = {Safari, Mojtaba and Yang, Xiaofeng and Chang, Chih-Wei and Qiu, Richard L J and Fatemi, Ali and Archambault, Louis},
	title = {Unsupervised MRI motion artifact disentanglement: introducing MAUDGAN},
	journal = {Physics in Medicine and Biology}
}

@inproceedings{spieker2024self,
	title={Self-supervised k-Space Regularization for Motion-Resolved Abdominal MRI Using Neural Implicit k-Space Representations},
	author={Spieker, Veronika and Eichhorn, Hannah and Stelter, Jonathan K and Huang, Wenqi and Braren, Rickmer F and Rueckert, Daniel and Sahli Costabal, Francisco and Hammernik, Kerstin and Prieto, Claudia and Karampinos, Dimitrios C and others},
	booktitle={International Conference on Medical Image Computing and Computer-Assisted Intervention},
	pages={614--624},
	year={2024},
	organization={Springer}
}

@inproceedings{eichhorn2024physics,
	title={Physics-informed deep learning for motion-corrected reconstruction of quantitative brain MRI},
	author={Eichhorn, Hannah and Spieker, Veronika and Hammernik, Kerstin and Saks, Elisa and Weiss, Kilian and Preibisch, Christine and Schnabel, Julia A},
	booktitle={International Conference on Medical Image Computing and Computer-Assisted Intervention},
	pages={562--571},
	year={2024},
	organization={Springer}
}

@article {Sommer416,
	author = {Sommer, K. and Saalbach, A. and Brosch, T. and Hall, C. and Cross, N.M. and Andre, J.B.},
	title = {Correction of Motion Artifacts Using a Multiscale Fully Convolutional Neural Network},
	volume = {41},
	number = {3},
	pages = {416--423},
	year = {2020},
	doi = {10.3174/ajnr.A6436},
	publisher = {American Journal of Neuroradiology},
	issn = {0195-6108},
	URL = {https://www.ajnr.org/content/41/3/416},
	eprint = {https://www.ajnr.org/content/41/3/416.full.pdf},
	journal = {American Journal of Neuroradiology}
}

@article{https://doi.org/10.1002/mrm.28719,
	author = {Lee, Jongyeon and Kim, Byungjai and Park, HyunWook},
	title = {MC2-Net: motion correction network for multi-contrast brain MRI},
	journal = {Magnetic Resonance in Medicine},
	volume = {86},
	number = {2},
	pages = {1077-1092},
	doi = {https://doi.org/10.1002/mrm.28719},
	url = {https://onlinelibrary.wiley.com/doi/abs/10.1002/mrm.28719},
	eprint = {https://onlinelibrary.wiley.com/doi/pdf/10.1002/mrm.28719},
	year = {2021}
}

@article{LIU202069_MRM,
	title = {Motion artifacts reduction in brain MRI by means of a deep residual network with densely connected multi-resolution blocks (DRN-DCMB)},
	journal = {Magnetic Resonance Imaging},
	volume = {71},
	pages = {69-79},
	year = {2020},
	issn = {0730-725X},
	doi = {https://doi.org/10.1016/j.mri.2020.05.002},
	url = {https://www.sciencedirect.com/science/article/pii/S0730725X20300175},
	author = {Junchi Liu and Mehmet Kocak and Mark Supanich and Jie Deng}
}

@article{ALMASNI2022119411,
	title = {Stacked U-Nets with self-assisted priors towards robust correction of rigid motion artifact in brain MRI},
	journal = {NeuroImage},
	volume = {259},
	pages = {119411},
	year = {2022},
	issn = {1053-8119},
	doi = {https://doi.org/10.1016/j.neuroimage.2022.119411},
	url = {https://www.sciencedirect.com/science/article/pii/S1053811922005286},
	author = {Mohammed A. Al-masni and Seul Lee and Jaeuk Yi and Sewook Kim and Sung-Min Gho and Young Hun Choi and Dong-Hyun Kim}
}

@article{https://doi.org/10.1002/mrm.70050,
	author = {Eichhorn, Hannah and Spieker, Veronika and Hammernik, Kerstin and Saks, Elisa and Felsner, Lina and Weiss, Kilian and Preibisch, Christine and Schnabel, Julia A.},
	title = {Motion-robust T2s quantification from low-resolution gradient echo brain MRI with physics-informed deep learning},
	journal = {Magnetic Resonance in Medicine},
	volume = {95},
	number = {1},
	pages = {346--362},
	year={2026},
	doi = {https://doi.org/10.1002/mrm.70050},
	url = {https://onlinelibrary.wiley.com/doi/abs/10.1002/mrm.70050},
	eprint = {https://onlinelibrary.wiley.com/doi/pdf/10.1002/mrm.70050}
}

@article{KIM2025109978,
	title = {Unsupervised learning for motion correction and assessment in brain magnetic resonance imaging using severity-based regularized cycle consistency},
	journal = {Engineering Applications of Artificial Intelligence},
	volume = {142},
	pages = {109978},
	year = {2025},
	issn = {0952-1976},
	doi = {https://doi.org/10.1016/j.engappai.2024.109978},
	url = {https://www.sciencedirect.com/science/article/pii/S0952197624021377},
	author = {Seuk Kim and Mohammed A. Al-masni and Seul Lee and Sunyoung Jung and Kyu-Jin Jung and Chuanjiang Cui and Sung-Min Gho and Young Hun Choi and Dong-Hyun Kim}
}

@article{https://doi.org/10.1002/mrm.24615,
	author = {Loktyushin, Alexander and Nickisch, Hannes and Pohmann, Rolf and Schölkopf, Bernhard},
	title = {Blind retrospective motion correction of MR images},
	journal = {Magnetic Resonance in Medicine},
	volume = {70},
	number = {6},
	pages = {1608-1618},
	doi = {https://doi.org/10.1002/mrm.24615},
	url = {https://onlinelibrary.wiley.com/doi/abs/10.1002/mrm.24615},
	eprint = {https://onlinelibrary.wiley.com/doi/pdf/10.1002/mrm.24615},
	year = {2013}
}

@article{https://doi.org/10.1002/mrm.26796,
	author = {Cordero-Grande, Lucilio and Hughes, Emer J. and Hutter, Jana and Price, Anthony N. and Hajnal, Joseph V.},
	title = {Three-dimensional motion corrected sensitivity encoding reconstruction for multi-shot multi-slice MRI: Application to neonatal brain imaging},
	journal = {Magnetic Resonance in Medicine},
	volume = {79},
	number = {3},
	pages = {1365-1376},
	doi = {https://doi.org/10.1002/mrm.26796},
	url = {https://onlinelibrary.wiley.com/doi/abs/10.1002/mrm.26796},
	eprint = {https://onlinelibrary.wiley.com/doi/pdf/10.1002/mrm.26796},
	year = {2018}
}

@inproceedings{
	wu2025moner,
	title={Moner: Motion Correction in Undersampled Radial {MRI} with Unsupervised Neural Representation},
	author={Qing Wu and Chenhe Du and Xuanyu Tian and Jingyi Yu and Yuyao Zhang and Hongjiang Wei},
	booktitle={The Thirteenth International Conference on Learning Representations},
	year={2025},
	url={https://openreview.net/forum?id=OdnqG1fYpo}
}

@ARTICLE{10639524,
	author={Dabrowski, Oscar and Falcone, Jean-Luc and Klauser, Antoine and Songeon, Julien and Kocher, Michel and Chopard, Bastien and Lazeyras, François and Courvoisier, Sébastien},
	journal={IEEE Transactions on Medical Imaging}, 
	title={SISMIK for Brain MRI: Deep-Learning-Based Motion Estimation and Model-Based Motion Correction in k-Space}, 
	year={2025},
	volume={44},
	number={1},
	pages={396-408},
	doi={10.1109/TMI.2024.3446450}}

@article{safari2025physicsinformeddeeplearningmodel,
	author = {Safari, Mojtaba and Wang, Shansong and Eidex, Zach and Qiu, Richard L. J. and Chang, Chih-Wei and Yu, David S. and Yang, Xiaofeng},
	title = {A physics-informed deep learning model for MRI brain motion correction},
	journal = {Medical Physics},
	volume = {52},
	number = {12},
	pages = {e70197},
	doi = {https://doi.org/10.1002/mp.70197},
	url = {https://aapm.onlinelibrary.wiley.com/doi/abs/10.1002/mp.70197},
	eprint = {https://aapm.onlinelibrary.wiley.com/doi/pdf/10.1002/mp.70197},
	year = {2025}
}

@inproceedings{10.1007/978-3-031-43999-5_28,
title={Dual domain motion artifacts correction for mr imaging under guidance of k-space uncertainty},
author={Wang, Jiazhen and Yang, Yizhe and Yang, Yan and Sun, Jian},
booktitle={International conference on medical image computing and computer-assisted intervention},
pages={293--302},
year={2023},
organization={Springer Nature Switzerland},
isbn={978-3-031-43999-5}
}

@InProceedings{10.1007/978-3-031-72104-5_37,
	  title={IM-MoCo: self-supervised MRI motion correction using motion-guided implicit neural representations},
	author={Al-Haj Hemidi, Ziad and Weihsbach, Christian and Heinrich, Mattias P},
	booktitle={International Conference on Medical Image Computing and Computer-Assisted Intervention},
	pages={382--392},
	year={2024},
	organization={Springer Nature Switzerland},
	isbn={978-3-031-72104-5}
}

@article{angella2025dimadiffusingmotionartifacts,
	title={DIMA: DIffusing Motion Artifacts for unsupervised correction in brain MRI images}, 
	author={Paolo Angella and Luca Balbi and Fabrizio Ferrando and Paolo Traverso and Rosario Varriale and Vito Paolo Pastore and Matteo Santacesaria},
	year={2025},
	eprint={2504.06767},
	archivePrefix={arXiv},
	primaryClass={eess.IV},
	url={https://arxiv.org/abs/2504.06767}, 
}

@ARTICLE{10375761,
	author={Oh, Gyutaek and Jung, Sukyoung and Lee, Jeong Eun and Ye, Jong Chul},
	journal={IEEE Transactions on Computational Imaging}, 
	title={Annealed Score-Based Diffusion Model for MR Motion Artifact Reduction}, 
	year={2024},
	volume={10},
	number={},
	pages={43-53},
	doi={10.1109/TCI.2023.3347917}}

@INPROCEEDINGS{9157636,
	author={He, Kaiming and Fan, Haoqi and Wu, Yuxin and Xie, Saining and Girshick, Ross},
	booktitle={2020 IEEE/CVF Conference on Computer Vision and Pattern Recognition (CVPR)}, 
	title={Momentum Contrast for Unsupervised Visual Representation Learning}, 
	year={2020},
	volume={},
	number={},
	pages={9726-9735},
	doi={10.1109/CVPR42600.2020.00975}}

@InProceedings{pmlr-v119-chen20j,
	title = 	 {A Simple Framework for Contrastive Learning of Visual Representations},
	author =       {Chen, Ting and Kornblith, Simon and Norouzi, Mohammad and Hinton, Geoffrey},
	booktitle = 	 {Proceedings of the 37th International Conference on Machine Learning},
	pages = 	 {1597--1607},
	year = 	 {2020},
	editor = 	 {III, Hal Daumé and Singh, Aarti},
	volume = 	 {119},
	series = 	 {Proceedings of Machine Learning Research},
	month = 	 {13--18 Jul},
	publisher =    {PMLR},
	url = 	 {https://proceedings.mlr.press/v119/chen20j.html}
}

@misc{chen2020improvedbaselinesmomentumcontrastive,
	title={Improved Baselines with Momentum Contrastive Learning}, 
	author={Xinlei Chen and Haoqi Fan and Ross Girshick and Kaiming He},
	year={2020},
	eprint={2003.04297},
	archivePrefix={arXiv},
	primaryClass={cs.CV},
	url={https://arxiv.org/abs/2003.04297}, 
}

@misc{rusak2025infonceidentifyinggaptheory,
	title={InfoNCE: Identifying the Gap Between Theory and Practice}, 
	author={Evgenia Rusak and Patrik Reizinger and Attila Juhos and Oliver Bringmann and Roland S. Zimmermann and Wieland Brendel},
	year={2025},
	eprint={2407.00143},
	archivePrefix={arXiv},
	primaryClass={cs.LG},
	url={https://arxiv.org/abs/2407.00143}, 
}

@ARTICLE{10643318,
	author={Azad, Reza and Aghdam, Ehsan Khodapanah and Rauland, Amelie and Jia, Yiwei and Avval, Atlas Haddadi and Bozorgpour, Afshin and Karimijafarbigloo, Sanaz and Cohen, Joseph Paul and Adeli, Ehsan and Merhof, Dorit},
	journal={IEEE Transactions on Pattern Analysis and Machine Intelligence}, 
	title={Medical Image Segmentation Review: The Success of U-Net}, 
	year={2024},
	volume={46},
	number={12},
	pages={10076-10095},
	doi={10.1109/TPAMI.2024.3435571}}

@InProceedings{He_2019_CVPR,
	author = {He, Jingwen and Dong, Chao and Qiao, Yu},
	title = {Modulating Image Restoration With Continual Levels via Adaptive Feature Modification Layers},
	booktitle = {Proceedings of the IEEE/CVF Conference on Computer Vision and Pattern Recognition (CVPR)},
	month = {June},
	year = {2019}
}

@ARTICLE{10738507,
	author={Wang, Longguang and Guo, Yulan and Wang, Yingqian and Dong, Xiaoyu and Xu, Qingyu and Yang, Jungang and An, Wei},
	journal={IEEE Transactions on Pattern Analysis and Machine Intelligence}, 
	title={Unsupervised Degradation Representation Learning for Unpaired Restoration of Images and Point Clouds}, 
	year={2025},
	volume={47},
	number={1},
	pages={1-18},
	doi={10.1109/TPAMI.2024.3471571}}

@InProceedings{Isola_2017_CVPR,
	author = {Isola, Phillip and Zhu, Jun-Yan and Zhou, Tinghui and Efros, Alexei A.},
	title = {Image-To-Image Translation With Conditional Adversarial Networks},
	booktitle = {Proceedings of the IEEE Conference on Computer Vision and Pattern Recognition (CVPR)},
	month = {July},
	year = {2017}
}

@article{miyato2018spectral,
	title={Spectral normalization for generative adversarial networks},
	author={Miyato, Takeru and Kataoka, Toshiki and Koyama, Masanori and Yoshida, Yuichi},
	journal={arXiv preprint arXiv:1802.05957},
	year={2018}
}

@article{VANESSEN201362,
	title = {The WU-Minn Human Connectome Project: An overview},
	journal = {NeuroImage},
	volume = {80},
	pages = {62-79},
	year = {2013},
	note = {Mapping the Connectome},
	issn = {1053-8119},
	doi = {https://doi.org/10.1016/j.neuroimage.2013.05.041},
	url = {https://www.sciencedirect.com/science/article/pii/S1053811913005351},
	author = {David C. {Van Essen} and Stephen M. Smith and Deanna M. Barch and Timothy E.J. Behrens and Essa Yacoub and Kamil Ugurbil}
}

@Article{Narai2022_mrART,
	author={N{\'a}rai, {\'A}d{\'a}m
	and Hermann, Petra
	and Auer, Tibor
	and Kemenczky, P{\'e}ter
	and Szalma, J{\'a}nos
	and Homolya, Istv{\'a}n
	and Somogyi, Eszter
	and Vakli, P{\'a}l
	and Weiss, B{\'e}la
	and Vidny{\'a}nszky, Zolt{\'a}n},
	title={Movement-related artefacts (MR-ART) dataset of matched motion-corrupted and clean structural MRI brain scans},
	journal={Scientific Data},
	year={2022},
	month={Oct},
	day={17},
	volume={9},
	number={1},
	pages={630},
	issn={2052-4463},
	doi={10.1038/s41597-022-01694-8},
	url={https://doi.org/10.1038/s41597-022-01694-8}
}

@article{perez_garcia_torchio_2021,
	title = {{TorchIO}: a {Python} library for efficient loading, preprocessing, augmentation and patch-based sampling of medical images in deep learning},
	journal = {Computer Methods and Programs in Biomedicine},
	pages = {106236},
	year = {2021},
	issn = {0169-2607},
	doi = {https://doi.org/10.1016/j.cmpb.2021.106236},
	url = {https://www.sciencedirect.com/science/article/pii/S0169260721003102},
	author = {P{\'e}rez-Garc{\'i}a, Fernando and Sparks, Rachel and Ourselin, S{\'e}bastien},
}

@article{https://doi.org/10.1002/hbm.24750,
	author = {Isensee, Fabian and Schell, Marianne and Pflueger, Irada and Brugnara, Gianluca and Bonekamp, David and Neuberger, Ulf and Wick, Antje and Schlemmer, Heinz-Peter and Heiland, Sabine and Wick, Wolfgang and Bendszus, Martin and Maier-Hein, Klaus H. and Kickingereder, Philipp},
	title = {Automated brain extraction of multisequence MRI using artificial neural networks},
	journal = {Human Brain Mapping},
	volume = {40},
	number = {17},
	pages = {4952-4964},
	doi = {https://doi.org/10.1002/hbm.24750},
	url = {https://onlinelibrary.wiley.com/doi/abs/10.1002/hbm.24750},
	eprint = {https://onlinelibrary.wiley.com/doi/pdf/10.1002/hbm.24750},
	year = {2019}
}

@article{FISCHL2012774,

	title = {FreeSurfer},
	journal = {NeuroImage},
	volume = {62},
	number = {2},
	pages = {774-781},
	year = {2012},
	issn = {1053-8119},
	doi = {https://doi.org/10.1016/j.neuroimage.2012.01.021},
	url = {https://www.sciencedirect.com/science/article/pii/S1053811912000389},
	author = {Bruce Fischl},
}

@article{WU2023107373,
	title = {Unsupervised dual-domain disentangled network for removal of rigid motion artifacts in MRI},
	journal = {Computers in Biology and Medicine},
	volume = {165},
	pages = {107373},
	year = {2023},
	issn = {0010-4825},
	doi = {https://doi.org/10.1016/j.compbiomed.2023.107373},
	url = {https://www.sciencedirect.com/science/article/pii/S0010482523008387},
	author = {Boya Wu and Caixia Li and Jiawei Zhang and Haoran Lai and Qianjin Feng and Meiyan Huang}
}

@article{SARKAR2025108684,
	title = {AutoDPS: An unsupervised diffusion model based method for multiple degradation removal in MRI},
	journal = {Computer Methods and Programs in Biomedicine},
	volume = {263},
	pages = {108684},
	year = {2025},
	issn = {0169-2607},
	doi = {https://doi.org/10.1016/j.cmpb.2025.108684},
	url = {https://www.sciencedirect.com/science/article/pii/S0169260725001014},
	author = {Arunima Sarkar and Ayantika Das and Keerthi Ram and Sriprabha Ramanarayanan and Suresh Emmanuel Joel and Mohanasankar Sivaprakasam}
}
\end{document}